\documentclass[lettersize,journal]{IEEEtran}
\usepackage{amsmath,amsfonts}
\usepackage{algorithmic}
\usepackage{array}
\usepackage[caption=false,font=normalsize,labelfont=sf,textfont=sf]{subfig}
\usepackage{textcomp}
\usepackage{multirow} 
\usepackage{stfloats}
\usepackage{booktabs}
\usepackage{url}
\usepackage{verbatim}
\usepackage{graphicx}
\usepackage{natbib}
\usepackage{hyperref}
\usepackage{algorithm}
\usepackage{algorithmic}

\usepackage{amssymb}
\usepackage{amsmath}
\usepackage{booktabs}
\usepackage{subcaption}
\usepackage{xcolor}       
\usepackage{multirow} 
\usepackage{bbding}
\usepackage{makecell} 
\usepackage{newfloat}
\usepackage{listings}
\usepackage{float}
\usepackage{cleveref}

\def\BibTeX{{\rm B\kern-.05em{\sc i\kern-.025em b}\kern-.08em
    T\kern-.1667em\lower.7ex\hbox{E}\kern-.125emX}}
\usepackage{balance}
\begin{document}

\title{Robust Prompt Tuning for Vision-Language Models with Mild Semantic Noise}
\author{Yansheng Gao, Yufei Zheng, Shengsheng Wang
\thanks{This work is supported by the National Natural Science Foundation of China (62376106).(Corresponding author: Shengsheng Wang.)

Shengsheng Wang, Yansheng Gao are with the College
of Computer Science and Technology, Key Laboratory of Symbolic Computation and Knowledge Engineering of Ministry of Education, Jilin University, Changchun 130012, China(e-mail: wss@jlu.edu.cn;
15615865653@163.com).
Yufei Zheng is with the College of Software, Key Laboratory of Symbolic Computation and Knowledge Engineering of Ministry of Education, Jilin University, Changchun, China(e-mail: yfzheng24@mails.jlu.edu.cn).
}
}

\markboth{Journal of \LaTeX\ Class Files,~Vol.~14, No.~8, August~2021}%
{Shell \MakeLowercase{\textit{et al.}}: A Sample Article Using IEEEtran.cls for IEEE Journals}

\maketitle

\begin{abstract}
Prompt tuning has shown promising results, but its robustness and generalization to unseen categories remain limited. Through our experiments, we demonstrate that the complete removal of semantic noise is a key factor restricting robustness. Existing methods typically suppress or filter out semantic noise in the prompt space, inadvertently hindering the model's robustness and its ability to generalize to unseen categories. To address this, we propose ANPrompt, a robust prompt tuning framework that actively incorporates weak semantic noise. By clustering weakly perturbed features into noise prompts and integrating them with learnable tokens in both the text and vision encoders, ANPrompt ensures controlled exposure to semantic variations. To enhance the visual pathway, we introduce the Noise-Resistant Visual Prompt Prototype (NRVPP), which stabilizes visual semantics under weak perturbations. Additionally, we propose a Weak Alignment Loss (WALoss) at the logits level to enforce consistency between clean and perturbed predictions, providing stable supervision. By combining weak semantic noise exposure with logits-based consistency, ANPrompt prevents overfitting to specific phrasings while preserving semantic integrity. Extensive experiments across 11 benchmarks, including base-to-new splits, show that ANPrompt consistently outperforms existing prompt tuning methods, offering superior robustness to semantic noise and improved generalization across tasks.

\end{abstract}


\begin{figure}[h]
\centering
\includegraphics[scale=0.35]{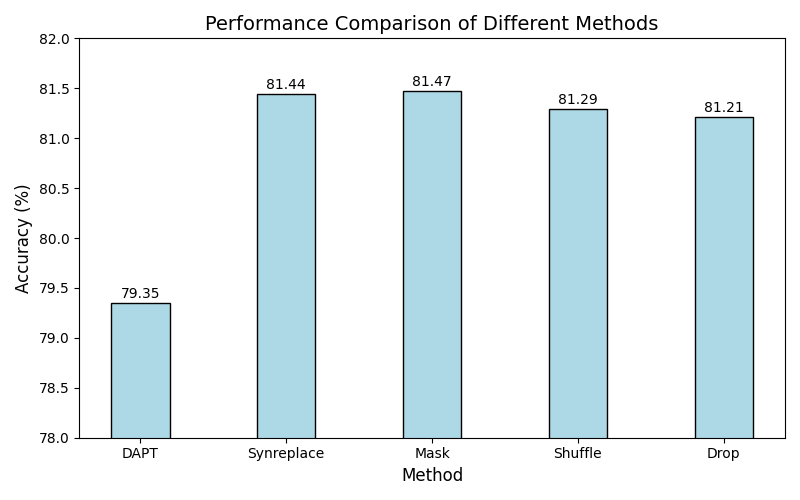}
\caption{This figure compares the accuracy of various methods, including DAPT, Synreplace, Mask, Shuffle, and Drop. The results show that incorporating weak semantic noise achieves superior performance, outperforming DAPT (79.35\%) and other baseline methods. This highlights the effectiveness of weak semantic noise injection in enhancing model robustness and generalization.}
\label{fig:intro_qua}
\end{figure}

\begin{figure*}[h]
\centering
\includegraphics[scale=0.6]{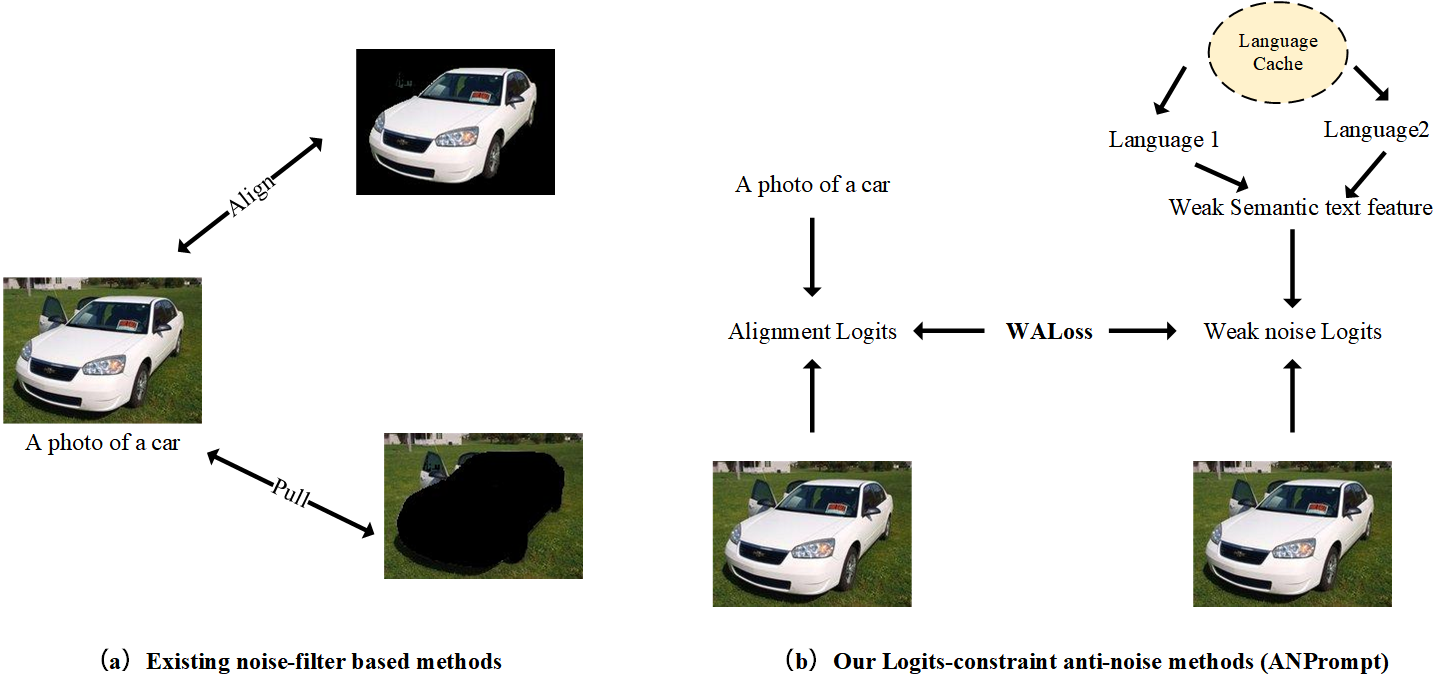}
\caption{Framework comparison with representative filtering-based methods such as DAPT and ArGue. 
While these approaches constrain frozen logits and risk overfitting under weak semantic perturbations, 
the proposed ANPrompt leverages weak semantic perturbations to generate soft logits and anti-noise prompts, 
thereby improving generalization.}
\label{fig:f1}
\end{figure*}

\begin{figure}[h]
\centering
\includegraphics[width=1.0\linewidth]{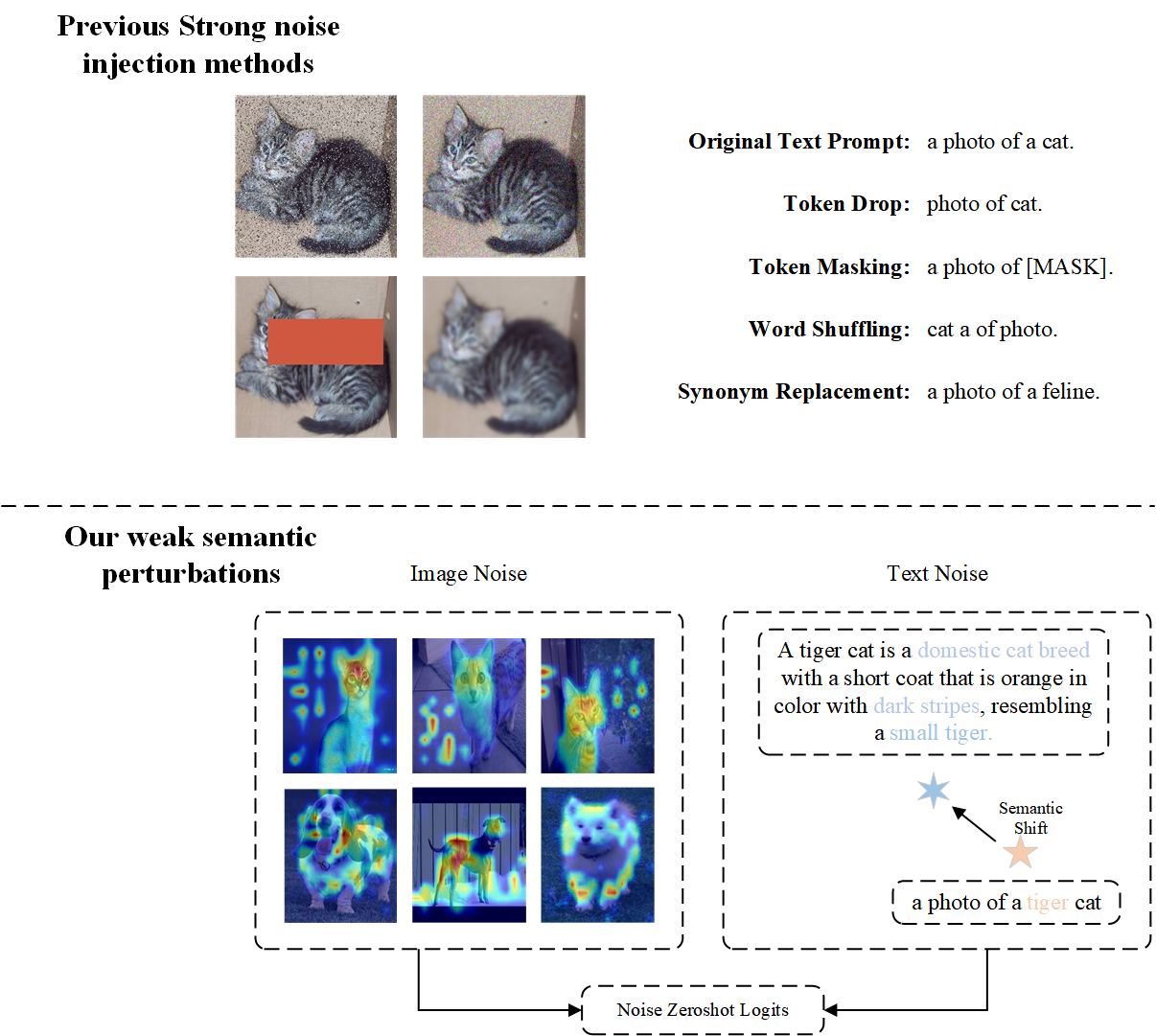}
\caption{Illustration of weak semantic perturbations in both image and text. 
Class activation maps (CAMs) may highlight irrelevant regions such as background or co-occurring objects, 
while subtle textual cues (e.g., ``dark stripes'') can shift class semantics from ``cat'' to ``tiger''. 
These examples demonstrate how weak semantic noise can mislead recognition and motivate the need for robust prompt learning.}
\label{fig:f2}
\end{figure}

\begin{figure*}[h]
\centering
\includegraphics[scale=0.26]{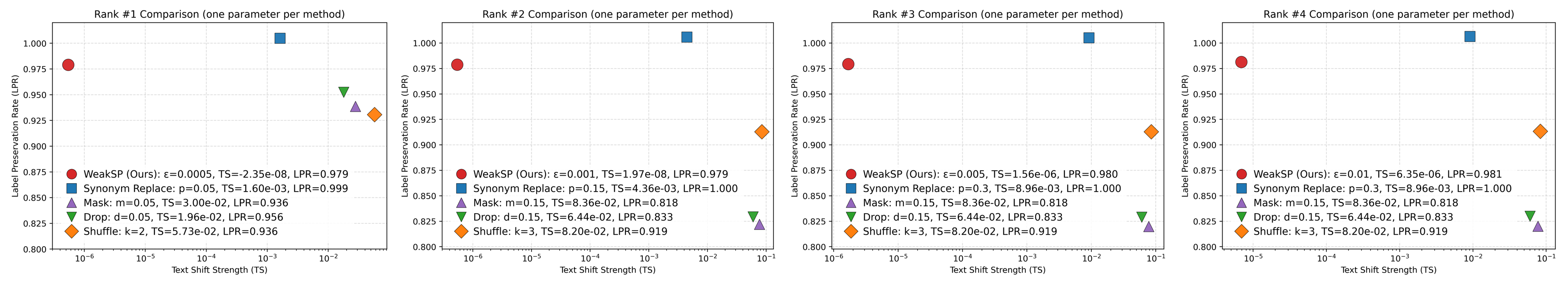}
\caption{Quantitative comparison of different noise strategies in terms of 
Text Shift (TS), Logit Preservation Rate (LPR), and Accuracy Shift (AS). 
The results show that strong perturbations (drop, mask, shuffle, synonym replacement) 
cause semantic distortion and performance degradation, whereas our weak semantic perturbations (WSP) 
maintain negligible shift, high prediction consistency, and nearly zero accuracy loss.}
\label{fig:f3}
\end{figure*}

\section{Introduction} \label{introduction}

Vision-language models (VLMs), such as CLIP, have demonstrated remarkable performance on open-vocabulary recognition tasks \cite{openworldauc, openvclip, xu2024llava}. However, adapting these large-scale models to downstream tasks is often infeasible due to the high computational cost of full fine-tuning. Prompt tuning \cite{awt, gallop, qnet} has emerged as an efficient alternative, updating only a small set of learnable tokens. Early approaches, such as CoOp \cite{coop}, tend to overfit to seen classes, leading to poor generalization to novel categories and sensitivity to semantic perturbations. Subsequent methods, including CoCoOp \cite{cocoop}, KgCoOp \cite{kgcoop}, MaPLe \cite{maple}, and MMRL \cite{mmrl}, introduce inductive biases or shared semantic spaces to enhance robustness, but they still struggle under noisy conditions.

Existing methods for handling noisy information in prompt tuning typically follow a filtering philosophy, where noise is suppressed or eliminated. Specifically, methods like DAPT \cite{dapt} and ArGue\cite{argue} align foreground features with textual descriptions while suppressing irrelevant background, as shown in Figure~\ref{fig:f1}(a). However, these techniques often risk discarding valuable semantic variations and struggle to adapt to subtle, real-world perturbations. Furthermore, we demonstrate that filtering or suppressing noise, as done by methods like DAPT (79.35\%) and Synreplace (81.44\%), often limits robustness. These approaches suppress irrelevant background noise, which not only discards valuable semantic information but also hinders the model’s generalization and robustness to diverse real-world perturbations.

In contrast, our method actively integrates weak semantic noise, resulting in a notable increase in performance. As shown in Figure~\ref{fig:intro_qua}, methods that filter or suppress noise show limited improvement, while our approach of incorporating weak semantic noise consistently outperforms these baseline methods. This emphasizes that actively incorporating weak semantic noise is more effective than filtering it, enabling the model to generalize better across unseen categories and enhancing its robustness to perturbations. As shown in Figure~\ref{fig:f1}(b), our method actively injects weak semantic noise, enhancing robustness by integrating it with constraint-based techniques, rather than suppressing it. This allows the model to better adapt to diverse inputs without sacrificing semantic integrity.

Noise injection has also been explored in the field of multimodal representation learning. Related works, such as \cite{noise1}, have proposed several text-based noise injection strategies, including strong perturbations like token dropping, masking, word shuffling, and synonym replacement, as illustrated in Figure~\ref{fig:f2}. While these strong perturbations introduce variation, they often distort the original semantics and degrade model generalization, as demonstrated in the top-left part of the figure. However, we find that such strong noise injection methods are not well-suited for the Prompt tuning domain. To address this issue, we propose a more gradual noise injection approach, introducing weak semantic perturbations, which apply subtle shifts in both the image and text domains. By incorporating a logits constraint, shown in the bottom portion of Figure~\ref{fig:f2}, our method ensures these perturbations preserve class identity while enhancing robustness. The stable Noise ZeroShot Logits generated by this approach help the model generalize better to unseen variations without the disruptive effects of strong noise perturbations.

To substantiate this, we run a controlled comparison under the base-to-new setting, as shown in Figure~\ref{fig:f3}. Strong perturbations severely distort semantics: token dropping yields $\mathrm{TS}=0.0196$ and $\mathrm{AS}=0.618$, masking further increases distortion with $\mathrm{TS}=0.0300$ and $\mathrm{AS}=0.923$, and word shuffling reaches $\mathrm{TS}=0.0573$ with $\mathrm{AS}=0.438$. Even synonym replacement introduces instability ($\mathrm{TS}=0.0090$, $\mathrm{AS}=0.00013$). In contrast, weak semantic noise preserves class identity, exhibiting negligible text shift ($\mathrm{TS}=6.3\times 10^{-6}$ at $\epsilon=0.01$), high prediction consistency ($\mathrm{LPR}\approx0.98$), and nearly zero accuracy degradation. These results suggest that \emph{controlled weak noise exposure} can enhance robustness without sacrificing semantic fidelity.

Building on these insights, we propose \textbf{ANPrompt}, a prompt-tuning framework that actively leverages weak semantic noise rather than suppressing it. Specifically, ANPrompt clusters weak noise features into representative prompts and integrates them with learnable tokens in both text and vision encoders. As shown in Figure~\ref{fig:f1}(b), our approach contrasts with previous methods that focus on noise suppression. By actively incorporating weak semantic noise, the model improves its robustness to unseen categories, while effectively enhancing generalization and maintaining semantic consistency. To further capture noise-aware visual semantics, we introduce a Noise-Resistant Visual Prompt Prototype (NRVPP), which provides stable visual representations under weak perturbations. This controlled noise exposure regularizes prompt representations, mitigates overfitting to specific textual expressions, and promotes learning of more robust class-level semantics. Furthermore, we introduce a \emph{Weak Alignment Loss (WALoss)} that enforces consistency between clean and perturbed predictions. Unlike feature-level perturbations, which may disrupt intermediate representations and lead to instability, enforcing consistency at the \emph{logits} level provides a more stable supervision signal. This directly regularizes the decision boundary and improves robustness without compromising the model's discriminative capacity. The benefits of WALoss align with our overall approach of controlled noise exposure, further enhancing the model's ability to adapt to real-world variations without losing accuracy or generalization.

Our main contributions are summarized as follows:
\begin{itemize}
\item We propose \textbf{ANPrompt}, a prompt tuning framework that improves robustness by incorporating weak semantic noise into both prompt tokens and logits.
\item We introduce a \textbf{Weak Alignment Loss (WALoss)} that enforces prediction consistency at the logits level, providing stable supervision and stronger generalization compared to feature-level perturbations.
\item We exploit \textbf{weak semantic noise} as a source of natural linguistic diversity, preserving class identity while preventing overfitting to specific phrasings, thereby enhancing base-to-new generalization compared with strong noise strategies.
\item We conduct extensive experiments on 11 datasets and base-to-new splits, demonstrating that ANPrompt achieves state-of-the-art robustness and generalization compared with existing prompt tuning methods.
\end{itemize}

\section{Related Works}
\label{relatedwork}
\subsection{Vision-Language Models}

Pretrained vision-language models (VLMs)\cite{l1, l2, l3, l4, l5} such as CLIP~\cite{clip} and ALIGN~\cite{align} achieve strong open-vocabulary and cross-modal capabilities through contrastive learning on large-scale image-text pairs. Building on this foundation, models like BLIP~\cite{blip}, FILIP~\cite{filip}, Florence~\cite{florence}, and LiT~\cite{lit} enhance fine-grained alignment and data efficiency via retrieval augmentation\cite{li2024freestyleret}, self-supervised learning, and stronger backbones. NLIP~\cite{NLIP} further improves robustness by introducing noise-harmonization based on alignment uncertainty and noise-completion via concept-guided caption generation. These advances enable effective zero-shot and few-shot transfer for tasks such as retrieval, classification, and grounding. However, adapting such large models to downstream tasks \cite{li2025beyond, li2023weakly, pcl,hipl,sod} remains costly due to domain shifts and the high overhead of full fine-tuning. To address this, parameter-efficient tuning methods—such as adapter-based approaches~\cite{clipadapter, mma, metaadapter} and prompt learning~\cite{protext, cpr, pmp}—have been proposed to enable lightweight adaptation without retraining the entire model.

\subsection{Prompt Learning}

Prompt learning has become an efficient strategy to adapt VLMs without full fine-tuning. Early methods such as CoOp~\cite{coop} and CoCoOp~\cite{cocoop} explore soft and image-conditioned prompts to enhance generalization. Subsequent works improve robustness and transferability through multimodal regularization (MaPLe~\cite{maple}, PromptSRC~\cite{promptsrc}), gradient alignment (ProGrad~\cite{prograd}), hierarchical and knowledge-guided prompts (HPT~\cite{hpt}, TCP~\cite{tcp}), and distillation-based initialization (CLIP-kd~\cite{clipkd}, ComKD-CLIP~\cite{comkdclip}, PromptKD~\cite{promptkd}, CasPL~\cite{caspl}). Other methods focus on attribution-based prompting (ATPrompt~\cite{atprompt}), two-stage decoupled optimization (DPC~\cite{dpc}, DePT~\cite{dept}, 2SFS~\cite{2sfs}), or cross-modal adapters (MMA~\cite{mma}) for few-shot settings. Consistency learning (CoPrompt~\cite{coprompt}) and task-aware clustering (TAC~\cite{tac}) further enhance adaptability. However, these approaches largely target generalization and efficiency, with limited attention to robustness against weak semantic perturbations.

\subsection{Prompting under Weak Semantic Perturbation Noise}

Recent works have explored robust prompt learning under noisy conditions. Methods such as NLPrompt~\cite{nlprompt}, JoAPR~\cite{joapr}, and ArGue~\cite{argue} mitigate label noise through loss design, sample filtering, and LLM-generated attributes, while ClipCleaner~\cite{clipcleaner} leverages CLIP-based sample selection to decouple training from noise filtering. However, these approaches mainly focus on filtering noise, which risks discarding informative variations and often limits generalization. In contrast, we propose ANPrompt, which embraces weak semantic perturbations as supervision signals. By integrating perturbed prompts with clustering-based bias representations, ANPrompt enables robust and discriminative prompt learning without relying on brittle filtering heuristics.

\section{Method}
\label{method}

\subsection{Preliminaries}
\label{sec:Preliminaries}

Vision-language models (VLMs) such as CLIP~\cite{clip} learn a joint embedding space for images and texts through contrastive learning over large-scale image-text pairs. These models exhibit strong zero-shot generalization but are highly sensitive to semantic variations in either modality.

A typical VLM consists of a vision encoder and a text encoder. The vision encoder tokenizes an input image into patches, encodes them through Transformer layers, and outputs a class token representing global semantics. The text encoder processes class-dependent prompts (e.g., ``a photo of a [CLASS]''), encodes them with Transformers, and uses the [EOS] token as the prompt embedding.

The classification score for class $c$ is computed via cosine similarity:
\begin{equation}
    p(y = c \mid f) = \frac{\exp(\text{sim}(f, w_c)/\tau)}{\sum_{i=1}^C \exp(\text{sim}(f, w_i)/\tau)}
\end{equation}
where $f$ is the image feature, $w_c$ is the text feature for class $c$, $\tau$ is a learnable temperature, and $\text{sim}(\cdot)$ denotes cosine similarity.

\begin{figure*}[t]
\centering
\includegraphics[scale=0.47]{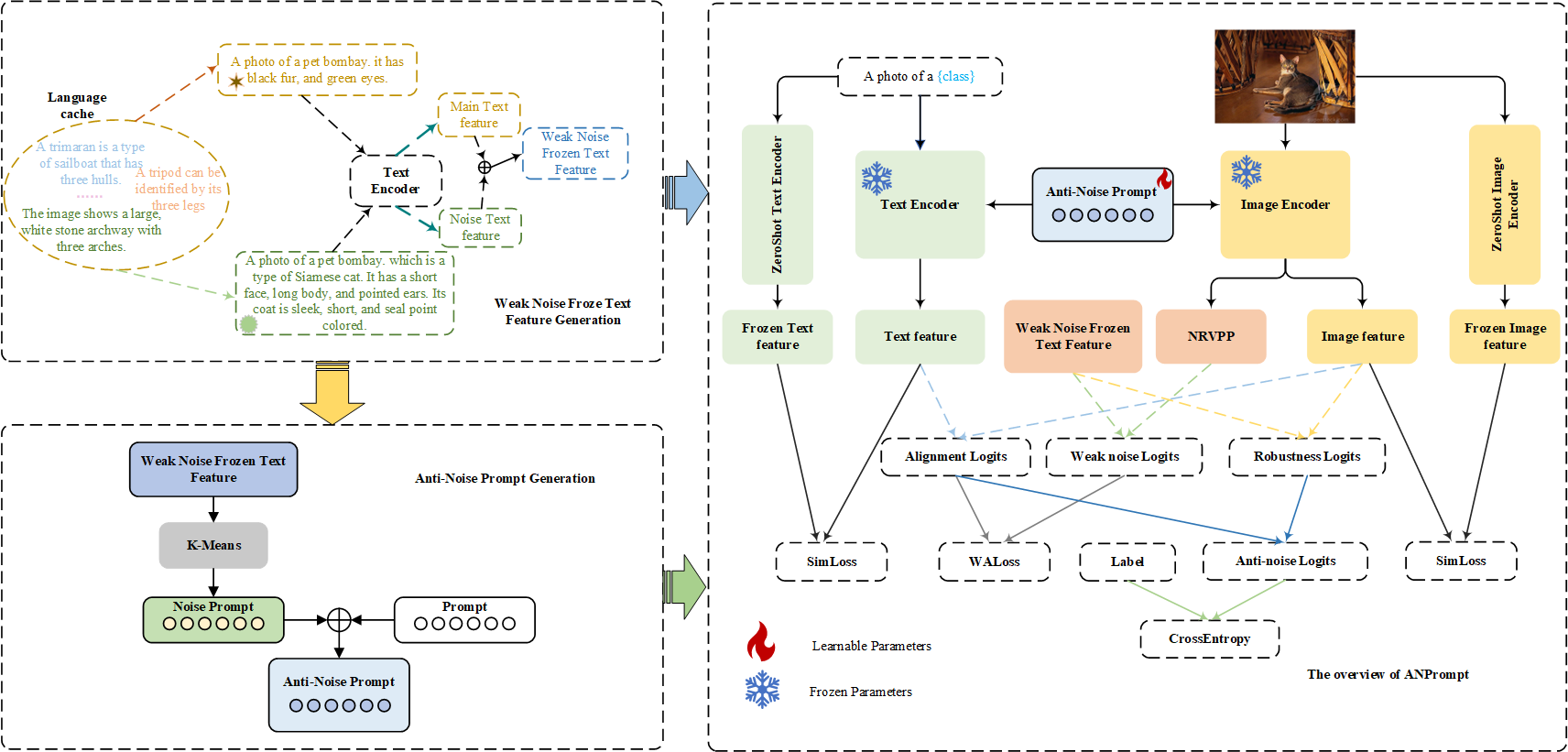}
\caption{Overview of ANPrompt. For each class, a main and a noise sentence are sampled to construct the Weak Noise Frozen Text Feature, which is clustered into noise prompts. These are combined with learnable tokens to generate Anti-noise Prompts, which are injected into the last two layers of both vision and text encoders. During encoding, we obtain Prompted Image/Text Features and the Noise-Resistant Visual Prompt Prototype (NRVPP) to compute four types of logits, supervised by WALoss and auxiliary losses to improve robustness against weak semantic perturbations.}
\label{fig:method}
\end{figure*}

\subsection{ANPrompt Framework}
\label{sec:ANPrompt}

We propose \textbf{ANPrompt}, a robust prompt tuning framework that explicitly leverages \textit{weak semantic noise} to enhance the robustness of VLMs. Specifically, ANPrompt simulates weak perturbations by fusing intra-class textual descriptions, generating \textit{Weak Noise Frozen Text Features}. These features are clustered to form noise prompts, which are added to learnable prompt tokens to construct \textit{Anti-noise Prompts}. The Anti-noise Prompts are injected into the last two transformer blocks of both vision and text encoders by prepending them as additional tokens, thereby exposing the model to weak semantic perturbations during prompt learning. After encoding, we obtain prompted image features, prompted text features, and the Noise-Resistant Visual Prompt Prototype (NRVPP), which are used to compute four types of logits: alignment, robustness, weak noise, and anti-noise logits. During training, ANPrompt combines the Weak Alignment Loss (WALoss) with cross-entropy and similarity losses to enhance robustness against weak semantic perturbations. Specifically, WALoss encourages the model to maintain prediction consistency between clean and perturbed inputs at the logits level, ensuring that the model remains stable even when exposed to subtle changes in the data. By incorporating weak semantic noise, ANPrompt regularizes the decision boundary, making it more resilient to small semantic shifts without sacrificing accuracy. Unlike previous methods, which primarily employ generic prompt regularization based on cosine similarity or feature-level perturbations, ANPrompt takes a more explicit approach by integrating weak semantic noise directly into prompt tokens. It enforces consistency at the logits level through WALoss, which provides a more stable supervision signal compared to traditional methods. This mechanism is distinctive in its ability to simultaneously preserve class identity and improve generalization, making ANPrompt a more effective solution for handling noise in vision-language models.

As illustrated in Figure~\ref{fig:method}, the ANPrompt framework proceeds through four sequential stages:  
(1) Weak Noise Frozen Text Feature Generation,  
(2) Anti-noise Prompt Construction,  
(3) Feature Encoding, and  
(4) Logit Computation.

\subsubsection{Weak Noise Frozen Text Feature Generation}
ANPrompt samples two intra-class descriptions: a \textit{main text} that encodes the core semantics and a \textit{noise text} that introduces subtle phrasing variations (e.g., attribute reordering). These two texts are encoded by the frozen CLIP text encoder into features $f_m$ and $f_n$, which are then fused as:

\begin{equation}
f_w = f_m + \alpha f_n ,
\end{equation}

where $\alpha$ is a scaling factor that controls the strength of the perturbation. The Weak Noise Frozen Text Feature $f_w$ combines contextual information from both the main and noise texts, capturing weak semantic perturbations that subtly shift semantics toward adjacent, but incorrect, classes.

The main and noise texts are randomly sampled from a diverse intra-class language cache. This approach effectively captures natural linguistic variations, approximating real-world weak perturbations. Future work may explore more controlled alternatives, such as paraphrasing or synonym replacement, to generate more fine-grained noise variations.

\subsubsection{Anti-noise Prompt Construction}

To map weak semantic perturbation information into the prompt space, ANPrompt applies \textbf{K-means clustering} on all $f_w$ to extract representative semantic centers, denoted as \textit{noise prompts}. These noise prompts are then added to the learnable prompt tokens $P_c$ to form intermediate anti-noise prompts $P_a$. Two modality-specific fully connected layers then transform $P_a$ into \textbf{Anti-noise Prompts} for the vision and text encoders, which are injected into their deeper layers.

\subsubsection{Feature Encoding}

After prompt injection, three types of features are obtained:
\begin{itemize}
    \item \textbf{Prompted Image Feature} $f_v$, encoded from the final image class token;
    \item \textbf{Prompted Text Feature} $f_t$, derived from the [EOS] token;
    \item \textbf{Noise-Resistant Visual Prompt Prototype (NRVPP)} $f_N$, obtained by mean-pooling the visual Anti-noise Prompts. This step helps capture weak semantic perturbations while suppressing high-frequency token-level variations, ensuring the preservation of meaningful visual information across perturbations.
\end{itemize}

\subsubsection{Logit Computation}

We compute four sets of logits:
\begin{itemize}
    \item \textbf{Alignment Logits} $\ell_A = f_v \cdot f_t^\top$, measuring the similarity between prompted image and text features;
    \item \textbf{Robustness Logits} $\ell_R = f_v \cdot f_w^\top$, assessing the model’s robustness to weak semantic noise in the text;
    \item \textbf{Weak Noise Logits} $\ell_W = f_N \cdot f_w^\top$, reflecting the alignment between noise-resistant visual features and weakly perturbed text features;
    \item \textbf{Anti-noise Logits} $\ell_{\text{final}} = \theta \ell_A + (1 - \theta)\ell_R$, integrating alignment and robustness for the final prediction.
\end{itemize}

After computing these four logits, we can compute the training losses: cross-entropy loss ($\mathcal{L}_{\text{CE}}$), weak alignment loss ($\mathcal{L}_{\text{WA}}$), and cosine similarity loss ($\mathcal{L}_{\text{sim}}$). Specifically:
\begin{itemize}
    \item The cross-entropy loss ($\mathcal{L}_{\text{CE}}$) is computed using the Alignment Logits $\ell_A$.
    \item The weak alignment loss ($\mathcal{L}_{\text{WA}}$) is computed based on the Weak Noise Logits $\ell_W$ and the Alignment Logits $\ell_A$.
    \item The cosine similarity loss ($\mathcal{L}_{\text{sim}}$) is calculated between the image and text embeddings to preserve semantic consistency across modalities.
\end{itemize}

The overall training loss is:
\begin{equation}
    \mathcal{L} = \mathcal{L}_{\text{CE}} + \lambda \mathcal{L}_{\text{sim}} + \gamma \mathcal{L}_{\text{WA}} ,
\end{equation}
where $\lambda$ controls the contribution of $\mathcal{L}_{\text{sim}}$, and $\gamma$ adaptively controls the effect of $\mathcal{L}_{\text{WA}}$. 

where $\mathcal{L}_{\text{CE}}$ is the cross-entropy loss, $\mathcal{L}_{\text{sim}}$ is the cosine similarity loss, and $\mathcal{L}_{\text{WA}}$ is the weak alignment loss. Here, $\mathcal{L}_{\text{sim}}$ is defined over both image and text embeddings to preserve semantic consistency across modalities. In addition, $K$ explicitly denotes the number of learnable prompt tokens, ensuring one-to-one integration with the clustered noise prompts.

\subsection{Weak Alignment Loss}
To enforce consistency under weak semantic perturbations, we introduce a KL divergence-based \textit{Weak Alignment Loss} ($\mathcal{L}_{\text{WA}}$) that minimizes the divergence between $\ell_W$ and $\ell_A$. Minimizing this divergence encourages predictions under perturbation to remain close to the clean alignment, penalizing semantic shifts while preserving structural consistency.

To compute $\ell_W$, the NRVPP $f_N$ is obtained by averaging visual prompt token features $f_v^o \in \mathbb{R}^{N \times C}$:
\begin{equation}
    f_N = \frac{1}{N} \sum_{i=1}^{N} f_v^{o(i)} ,
\end{equation}
which suppresses token-level variations and mitigates \textit{visual semantic distraction} caused by weak noise. The weak noise logits are then:
\begin{equation}
    \ell_W = f_N \cdot f_w^\top .
\end{equation}

The Weak Alignment Loss is defined as:
\begin{equation}
    \mathcal{L}_{\text{WA}} = \text{KL}(\text{softmax}(\ell_A) \,\|\, \text{softmax}(\ell_W)) ,
\end{equation}
We use $\gamma$ to control the variance-adaptive weight:
\begin{equation}
    \gamma = \frac{1}{\text{std}(\ell_R) \cdot |\ell_R| + \epsilon_0} ,
\end{equation}
where $\text{std}(\ell_R)$ is the standard deviation of $\ell_R$, $|\ell_R|$ is its number of elements, and $\epsilon_0$ is a small constant for numerical stability. This dynamic scaling adjusts the loss weight based on prediction dispersion, stabilizing optimization across varied confidence levels.

\subsection{Weak Semantic Perturbation Prompt Injection}
Finally, we formalize the injection mechFanism. The weak noise features $f_w \in \mathbb{R}^{N \times C}$ are clustered into $K$ noise prompts $P_w \in \mathbb{R}^{K \times C}$ using K-means, with $K$ matching the number of learnable prompt tokens $P_c \in \mathbb{R}^{K \times C}$. The anti-noise prompts are then defined as:
\begin{equation}
    P_a = P_c + \epsilon P_w ,
\end{equation}
where $\epsilon$ controls the perturbation strength. $P_a$ is projected into modality-specific spaces to obtain $P_a^{\text{image}}$ and $P_a^{\text{text}}$, which are injected into the input sequences:
\begin{equation}
\begin{aligned}
    &\{P_a^{\text{image}}, e_{\text{cls}}, e_1, \dots, e_M\}, \\
    &\{t_{\text{SOS}}, P_a^{\text{text}}, t_1, \dots, t_L, c_k, t_{\text{EOS}}\}.
\end{aligned}
\end{equation}

This injection enriches the prompt space with weakly perturbed cues, promoting robustness against subtle noise while maintaining structural alignment with the original prompt format.

\begin{table*}[h]
    \centering
    \caption{Comparison with state-of-the-art methods on base-to-novel generalization. The best accuracies are bold. We can see that our method achieves the best performance compared with other state-of-the-art methods.}
    \label{tab:basetonew}
    \footnotesize 
    \begin{tabular}{l c c c c @{\hspace{10pt}} c c c @{\hspace{10pt}} c c c @{\hspace{10pt}} c c c}
        \toprule
        \multirow{2}{*}{\thead{Method}} & \multirow{2}{*}{\thead{Ref.}}
        & \multicolumn{3}{c}{Average} 
        & \multicolumn{3}{c}{ImageNet} 
        & \multicolumn{3}{c}{Caltech101} 
        & \multicolumn{3}{c}{OxfordPets} \\
        \cmidrule(lr){3-5} \cmidrule(lr){6-8} \cmidrule(lr){9-11} \cmidrule(lr){12-14}
        & & Base & Novel & HM  & Base & Novel & HM & Base & Novel & HM & Base & Novel & HM \\
        \midrule
        CLIP & ICML22 & 69.34 & 74.22 & 71.70 & 72.43 & 68.14 & 70.22 & 96.84 & 94.00 & 95.40 & 91.17 & 97.26 & 94.12 \\
        CoOp & IJCV22 & 82.69 & 63.22 & 71.66 & 76.47 & 67.88 & 71.92 & 98.00 & 89.81 & 94.20 & 93.67 & 95.29 & 94.47 \\
        CoCoOp & CVPR22 & 80.47 & 71.69 & 75.83 & 77.60 & 70.75 & 74.00 & 97.96 & 93.81 & 95.84 & 95.20 & 97.69 & 96.43 \\
        MaPLe & CVPR23 & 82.28 & 75.14 & 78.55 & 76.66 & 70.54 & 73.47 & 97.74 & 94.36 & 96.02 & 95.43 & 97.76 & 96.58 \\
        PromptSRC & ICCV23 & 84.26 & 76.10 & 79.97 & 77.60 & 70.73 & 74.01 & 98.10 & 94.03 & 96.02 & 95.33 & 97.30 & 96.30 \\      
        CoPrompt & ICLR24& 84.00 & 77.23 & 80.48 & 77.67 & 71.27 & 74.33 & 98.27 & 94.90 & 96.55 & 95.87 & \textbf{98.10} & 96.87 \\
        TextRefiner & AAAI25 & 79.74 & 74.32 & 76.94 & 76.84 & 70.54 & 73.56 & 98.13 & 94.43 & 96.24 & 95.27 & 97.65 & 96.45 \\
        MMRL & CVPR25 & 85.68 & 77.16 & 81.20 & 77.90 & 71.30 & \textbf{74.45} & 98.97 & 94.50 & 96.68 & \textbf{95.90} & 97.60 & 96.74 \\
        SurPL-G & ICML25 & \textbf{86.37} & 76.32 & 81.03 & \textbf{78.74} & 70.49 & 74.39 & 98.77 & 95.16 & 96.93 & 96.37 & 97.41 & \textbf{96.89} \\
        LwEIB & IJCV25 & 84.45 & \textbf{78.21} & 81.21 & 76.64 & \textbf{71.64} & 74.06 & 98.47 & \textbf{95.47} & \textbf{96.95} & 95.70 & 97.40 & 96.54 \\
        BIP-D & TPAMI25 & 85.95 & 75.79 & 80.55 & 78.27 & 69.90 & 73.85 & 98.77 & 94.47 & 96.57 & 95.30 & 97.37 & 96.32 \\
        \midrule
        \textbf{ANPrompt} & & 86.15 & 77.70 & \textbf{81.70} & 77.83 & 71.17 & 74.35 & \textbf{98.97} & 94.73 & 96.80 & 95.73 & 97.17 & 96.44 \\
        \bottomrule
    \end{tabular}
    
    \begin{tabular}{l c c c c @{\hspace{10pt}} c c c @{\hspace{10pt}} c c c @{\hspace{10pt}} c c c}
        \toprule
        \multirow{2}{*}{\thead{Method}} & \multirow{2}{*}{\thead{Ref.}}
        & \multicolumn{3}{c}{StandfordCars} 
        & \multicolumn{3}{c}{Flowers102} 
        & \multicolumn{3}{c}{Food101} 
        & \multicolumn{3}{c}{FGVCAircraft} \\
        \cmidrule(lr){3-5} \cmidrule(lr){6-8} \cmidrule(lr){9-11} \cmidrule(lr){12-14}
        & & Base & Novel & HM  & Base & Novel & HM & Base & Novel & HM & Base & Novel & HM \\
        \midrule
        CLIP & ICML22 & 63.37 & 74.89 & 68.65 & 72.08 & \textbf{77.80} & 74.83 & 90.10 & 91.22 & 90.66 & 27.19 & 36.29 & 31.09 \\
        CoOp & IJCV22 & 78.12 & 60.40 & 68.13 & 97.60 & 59.67 & 74.06 & 88.33 & 82.26 & 85.19 & 40.44 & 22.30 & 28.75 \\
        CoCoOp & CVPR22 & 70.49 & 73.59 & 72.01 & 94.87 & 71.75 & 81.71 & 90.70 & 91.29 & 90.99 & 33.41 & 23.71 & 27.74 \\
        MaPLe & CVPR23 & 72.94 & 74.00 & 73.47 & 95.92 & 72.46 & 82.56 & 90.71 & 92.05 & 91.38 & 37.44 & 35.61 & 36.50 \\
        PromptSRC & ICCV23 & 78.27 & 74.97 & 76.58 & 98.07 & 76.50 & 85.95 & 90.67 & 91.53 & 91.10 & 42.73 & 37.87 & 40.15 \\       
        CoPrompt & ICLR24& 76.97 & 74.40 & 75.66 & 97.27 & 76.60 & 85.71 & 90.73 & \textbf{92.07} & \textbf{91.40} & 40.20 & 39.33 & 39.76 \\
        TextRefiner & AAAI25 & 71.40 & 70.90 & 71.15 & 95.92 & 74.33 & 83.76 & 90.88 & 91.43 & 91.15& 35.35 & 35.87 & 35.61 \\
        MMRL & CVPR25 & 81.30 & \textbf{75.07} & 78.06 & \textbf{98.97} & 77.27 & \textbf{86.78} & 90.57 & 91.50 & 91.03 & 46.30 & 37.03 & 41.15 \\
        SurPL-G & ICML25 & 83.57 & 72.77 & 77.80 & 98.90 & 72.88 & 83.92 & \textbf{90.92} & 91.81 & 91.36 & 49.20 & 36.93 & 42.19 \\
        LwEIB & IJCV25 & 80.07 & 74.01 & 76.92 & 97.53 & 77.50 & 86.37 & 90.63 & 91.73 & 91.18 & 45.11 & \textbf{42.60} & \textbf{43.82} \\
        BIP-D & TPAMI25 & 83.27 & 73.00 & 77.80 & 98.30 & 75.90 & 85.66 & 90.73 & 91.23 & 90.98 & 48.83 & 35.90 & 41.38 \\
        \midrule
        \textbf{ANPrompt} & & \textbf{83.57} & 74.63 & \textbf{78.85} & 98.60 & 77.30 & 86.66 & 90.63 & 91.50 & 91.06 & \textbf{49.67} & 36.60 & 42.14 \\
        \bottomrule
    \end{tabular}

    \begin{tabular}{l c c c c @{\hspace{10pt}} c c c @{\hspace{10pt}} c c c @{\hspace{10pt}} c c c}
        \toprule
        \multirow{2}{*}{\thead{Method}} & \multirow{2}{*}{\thead{Ref.}}
        & \multicolumn{3}{c}{SUN397} 
        & \multicolumn{3}{c}{DTD} 
        & \multicolumn{3}{c}{EuroSAT} 
        & \multicolumn{3}{c}{UCF101} \\
        \cmidrule(lr){3-5} \cmidrule(lr){6-8} \cmidrule(lr){9-11} \cmidrule(lr){12-14}
        & & Base & Novel & HM  & Base & Novel & HM & Base & Novel & HM & Base & Novel & HM \\
        \midrule
        CLIP & ICML22 & 69.36 & 75.35 & 72.23 & 53.24 & 59.90 & 56.37 & 56.48 & 64.05 & 60.03 & 70.53 & 77.50 & 73.85 \\
        CoOp & IJCV22 & 80.60 & 65.89 & 72.51 & 79.44 & 41.18 & 54.24 & 92.19 & 54.74 & 68.69 & 84.69 & 56.05 & 67.46 \\
        CoCoOp & CVPR22 & 79.74 & 76.86 & 78.27 & 77.01 & 56.00 & 64.85 & 87.49 & 60.04 & 71.21 & 82.33 & 73.45 & 77.64 \\
        MaPLe & CVPR23 & 80.82 & 78.70 & 79.75 & 80.36 & 59.18 & 68.16 & 94.07 & 73.23 & 82.35 & 83.00 & 78.66 & 80.77 \\
        PromptSRC & ICCV23 & 82.67 & 78.47 & 80.52 & 83.37 & 62.97 & 71.75 & 92.90 & 73.90 & 82.30 & 87.10 & 78.80 & 82.74 \\       
        CoPrompt & ICLR24 & 82.63 & \textbf{80.03} & \textbf{81.31} & 83.13 & 64.73 & 72.79 & 94.60 & 78.57 & 85.84 & 86.90 & 79.57 & 83.07 \\
        TextRefiner & AAAI25 & 80.96 & 76.49 & 78.66 & 75.35 & 58.09 & 65.60 & 74.57 & 75.82 & 73.68 & 82.52 & 75.01 & 78.59 \\
        MMRL & CVPR25 & 83.20 & 79.30 & 81.20 & 85.67 & 65.00 & 73.82 & 95.60 & 80.17 & 87.21 & 88.10 & 80.07 & 83.39 \\
        SurPL-G & ICML25 & \textbf{83.43} & 78.96 & 81.13 & \textbf{86.07} & 62.04 & 72.11 & 94.63 & 81.33 & 87.48 & \textbf{89.44} & 79.74 & \textbf{84.31} \\
        LwEIB & IJCV25 & 81.10 & 79.80 & 80.44 & 82.87 & \textbf{67.83} & \textbf{74.60} & 95.00 & 80.01 & 86.86 & 85.73 & \textbf{82.37} & 84.02 \\
        BIP-D & TPAMI25 & 83.07 & 79.13 & 81.05 & 85.43 & 58.77 &69.63 & \textbf{95.70} & 78.00 & 85.95 & 87.77 & 80.00 & 83.70 \\
        \midrule
        \textbf{ANPrompt} & & 83.07 & 79.07 & 81.02 & 85.20 & 65.10 & 73.80 & 95.53 & \textbf{87.33} & \textbf{91.21} & 88.80 & 80.07 & 84.21 \\
        \bottomrule
    \end{tabular}
\end{table*}

\begin{table*}[h]
    \centering
    \caption{Results from cross-dataset evaluation. ANPrompt provides the highest average accuracy.}
    \label{tab:crossdataset}
    \footnotesize
    \setlength{\tabcolsep}{1pt} 
    \begin{tabular}{lcccccccccccccc}
        \toprule
        \multirow{2}{*}{} & Source &  &  &  &  &  & Target  \\
        \cmidrule{3-13}
        & ImageNet & Caltech & Pets & Cars & Flowers & Food & Aircraft & SUN397 & DTD & EuroSat & UCF101 & Average \\
        \midrule
        CoOp & 71.51 & 93.70 & 89.14 & 64.51 & 68.71 & 85.30 & 18.47 & 64.15 & 41.90 & 46.39 & 66.55 & 63.88 \\
        CoCoOp & 71.02 & 94.43 & 90.14 & 65.32 & 71.88 & 86.06 & 22.94 & 67.36 & 45.73 & 45.37 & 68.21 & 65.74 \\
        MaPLe & 70.72 & 93.53 & 90.55 & 65.57 & 72.23 & 86.20 & 24.74 & 67.01 & 46.49 & 48.06 & 68.06 & 66.10 \\
        PromptSRC & 71.27 & 93.60 & 90.25 & 65.70 & 70.25 & 86.15 & 23.90 & 67.10 & 46.87 & 45.50 & 68.75 & 65.81 \\
        CoPrompt & 70.80 & 94.50 & 90.75 & 65.67 & 72.70 & \textbf{86.43} & 20.40 & 67.57 & 47.07 & 51.90 & 69.73 & 67.00 \\
        TAC & 72.77 & 94.53 & 90.67 & 65.30 & 72.20 & 85.83 & 23.53 & \textbf{67.63} & 47.57 & 48.07 & 70.00 & 66.53 \\
        SurPL-G & \textbf{73.33} & 93.73 & 90.16 & 64.67 & 70.53 & 85.52 & 24.80 & 67.43 & \textbf{48.54} & \textbf{52.85} & 67.86 & 66.61 \\
        BIP &70.83 & 93.93 & 90.13 & 65.53 & 71.43 & 86.27 & \textbf{26.47} & 67.23 & 46.00 & 48.90 & \textbf{69.83} & 66.57 \\
        \midrule
        \textbf{ANPrompt} & 71.13 & \textbf{94.70} & \textbf{91.00} & \textbf{66.00} & \textbf{72.87} & 86.23 & 26.10 & 67.47 & 45.83 & 51.97 & 69.27 & \textbf{67.14} \\
        \bottomrule
    \end{tabular}
\end{table*}
\section{EXPERIMENT}
\label{experiment}
\subsection{Dataset} 

In the tasks of Base-to-New Generalization and Cross-Dataset Evaluation, we followed the experimental setup of CoOp~\cite{coop}, and evaluated on 11 public datasets, including ImageNet, Caltech~\cite{caltech101}, OxfordPets~\cite{oxfordpets}, Flowers~\cite{Flowers102}, Food101~\cite{food101}, StanfordCars~\cite{StanfordCars}, FGVCAircraft~\cite{fgvcaircraft}, EuroSAT~\cite{Eurosat}, UCF101~\cite{ucf101}, DTD~\cite{dtd} and SUN397~\cite{sun}. We also use ImageNet~\cite{imagenet} as the source dataset, and use its four variants ImageNetV2~\cite{imagenetv2}, ImageNet-Sketch~\cite{imagenets}, ImageNet-A~\cite{imageneta} and ImageNet-R~\cite{imagenetr} as the target dataset to evaluate the domain generalization task.

\subsection{Implementation Details} 

We set the learning rate to 0.001 and adopt the Adam optimizer to train ANPrompt. The model is trained for 10 epochs with a batch size of 4, using ViT-B/16 as the vision backbone. For the LLM cache, we incorporate prompts from CoPrompt~\cite{coprompt} and HPT~\cite{hpt}. And we set the $\theta$ and $\alpha$ with 0.7 and 0.001. We inject the prompt into the layer 6 to 12. All experiments are conducted on a single NVIDIA RTX 3090 Ti GPU. Reported results are averaged over three random seeds.

\subsection{Base-to-New Generalization} 
In this experiment, we evaluate ANPrompt and benchmark its performance against several existing methods, such as CLIP~\cite{clip}, CoOp~\cite{coop}, CoCoOp~\cite{cocoop}, MaPLe~\cite{maple}, PromptSRC~\cite{promptsrc}, CoPrompt~\cite{coprompt}, TextRefiner~\cite{textrefiner}, MMRL~\cite{mmrl}, SurPL-G~\cite{surpl}, LwEIB~\cite{lweib} and BIP-D~\cite{bip}.

\tablename~\ref{tab:basetonew} shows the performance of ANPrompt and other comparison methods on B2N generalization tasks. Compared with the most advanced LwEIB method, ANPrompt improves the base class by 1.7\% and the harmonic average by 0.49\%.  It is worth noting that compared with BIP-D, ANPrompt achieves significant performance improvement on EuroSAT dataset: 9.33\% on new class and 5.26\% on harmonic average. By transforming weak semantic noise into constructive training signals, ANPrompt achieves performance improvement by fusing complementary semantic perturbation sentences.

\subsection{Cross-Dataset Generalization} 

\tablename~\ref{tab:crossdataset} shows the performance comparison between our ANPrompt and existing methods in cross-dataset evaluation. On the ImageNet source dataset, ANPrompt shows comparable performance to other competitive methods. On the target domain dataset, ANPrompt achieved the best performance on four of the ten datasets, showing excellent generalization ability, especially on fine-grained datasets, such as FGVCAircraft and EuroSAT. Specifically, with an average of 10 datasets, ANPrompt achieved the highest accuracy rate of 67.14\%. Meanwhile, compared with the suboptimal method CoPrompt, ANPrompt achieved 0.14\% performance improvement. Different from the existing methods, we use the anti-noise visual prompt prototype to align the weak noise frozen text features to generate soft, context-aware supervision signals, thus improving cross-domain performance.

\begin{table}[ht]
\centering
\caption{Results on domain generalization task.}
\label{tab:doamingeneralization}
\footnotesize
\setlength{\tabcolsep}{3pt} 
\begin{tabular}{lcccccccc}
    \toprule
    \multirow{2}{*}{} & Source &  &  & Target \\
    \cmidrule{3-7}
    & ImageNet & -V2 & -Sk & -A & -R & Average \\
    \midrule
    CLIP & 66.73 & 60.83 & 46.15 & 47.77 & 73.96 & 57.18 \\
    CoOp & 71.51 & 64.20 & 47.99 & 49.71 & 75.21 & 59.28 \\ 
    CoCoOp &  71.02 & 64.07 & 48.75 & 50.63 & 76.18 & 59.90 \\ 
    MaPLe & 70.72 & 64.07 & 49.15 & 50.90 & 76.98 & 60.26 \\ 
    PromptSRC & 71.27 & 64.35 & 49.55 & 50.90 & \textbf{77.80} & 60.63 \\ 
    CoPrompt & 70.80 & 64.25 & 49.43 & 50.50 & 77.51 & 60.42 \\ 
    MMRL & \textbf{72.03} & 64.47 & 49.17 & \textbf{51.20} & 77.53 & \textbf{60.59} \\
    HiCroPL & 71.22 & 64.33 & 49.47 & 50.79 & 77.15 & 60.44 \\
    BIP & 70.80 & 64.40 & \textbf{50.00} & 50.30 & 77.30 & 60.50 \\
    \midrule
    \textbf{ANPrompt} & 71.13 & \textbf{64.63} & 49.13 & 50.37 & 77.47 & 60.40 \\
    \bottomrule
\end{tabular}
\end{table}

\subsection{Domain Generalization}

We compare the performance of our ANPrompt with that of nine previous methods on cross-domain data sets. As shown in \tablename~\ref{tab:doamingeneralization}, the average performance of ANPrompt on four data sets exceeds that of most comparison methods, showing its strong cross-domain generalization ability. Specifically, ANPrompt performs best on Imagenet-V2 dataset, and it also shows considerable performance on other target domain datasets.

\subsection{Few-shot Learning}
\begin{figure*}[t]
    \centering
    \includegraphics[scale=0.4]{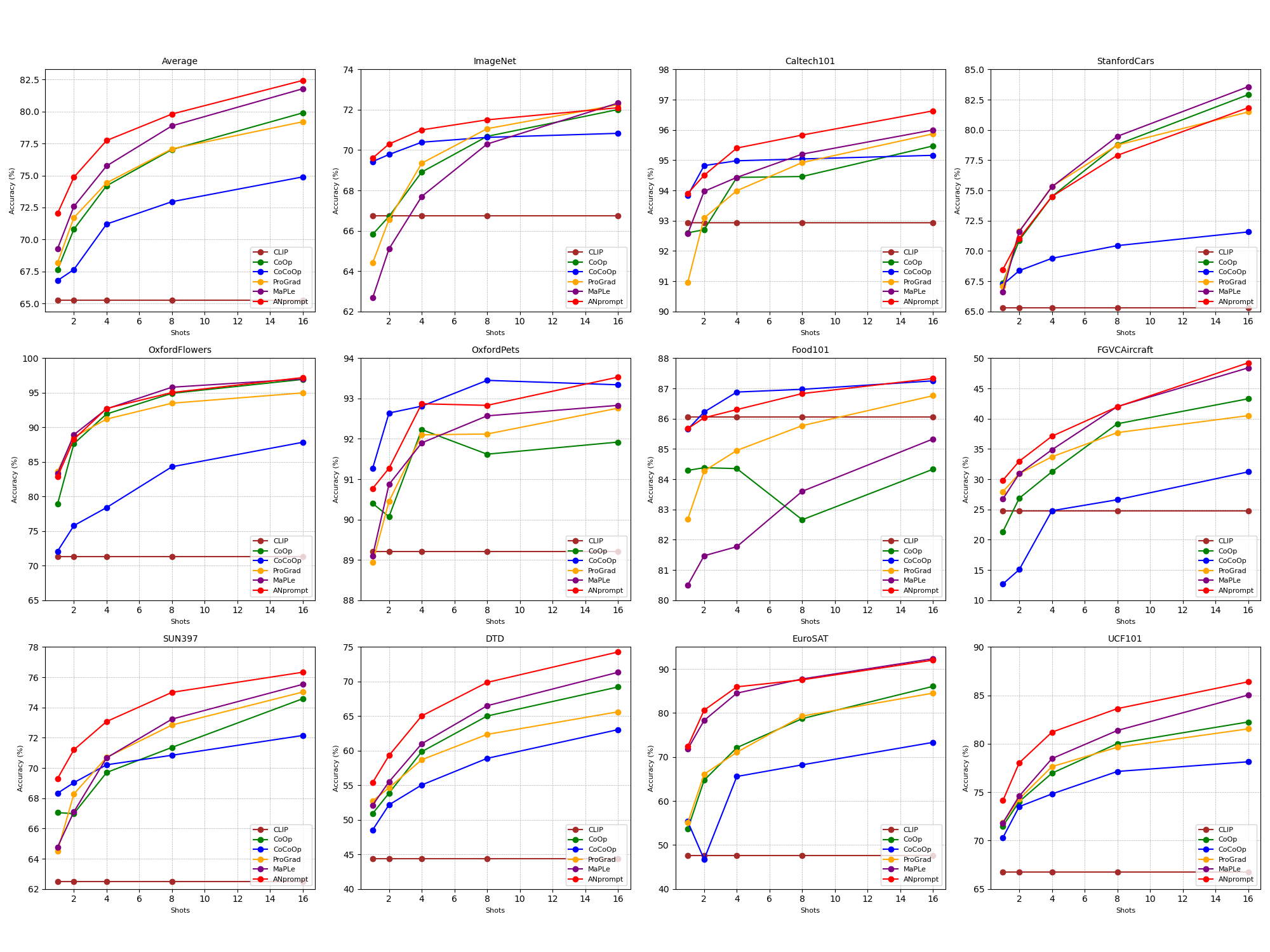}
    \caption{Few-shot classification accuracy across multiple datasets under varying shots. ANPrompt consistently outperforms existing methods.}
    \label{fig:fewshot}
\end{figure*}

We evaluate the discriminating  ability of the model under limited supervision by changing the number of shots in each category (K = 1, 2, 4, 8, 16). The evaluation includes five baseline methods and ANPrompt proposed by us, such as CLIP, CoOp, CoCoOp, Prograd and Maple. As shown in Figure~\ref{fig:fewshot}, ANPrompt is always superior to the comparison method in all lens settings, which shows the robustness of ANPrompt. Specifically, compared with the suboptimal method Maple, ANPrompt achieved an average performance improvement of 2.76\%, 2.29\%, 1.99\%, 0.92\% and 0.65\% on 1, 2, 4, 8 and 16 shots on 11 data sets, respectively. In terms of a single data set, ANPrompt showed the best accuracy on 8 of 11 datasets. For example, in 16shots, ANPrompt achieved 74.27\% value on DTD dataset, which was 2.94\% higher than Maple. Compared with Maple, UCF101 is improved by 1.37\%. In addition, ANPrompt's performance on other datasets is still competitive. This shows that ANPrompt effectively enhances the robustness under weak semantic disturbance by injecting weak semantic disturbance into the alignment process and prompt representation.

We compare the performance of our ANPrompt with that of six previous methods on cross-domain data sets. As shown in \tablename~\ref{tab:doamingeneralization}, the average performance of ANPrompt on four data sets exceeds that of most comparison methods, showing its strong cross-domain generalization ability. Specifically, ANPrompt performs best on Imagenet-V2 dataset, and it also shows considerable performance on other target domain datasets.

\subsection{Ablation Study} 

\subsubsection{Component ablation}
We conduct an ablation study to evaluate the contribution of each component in ANPrompt, including the Weak noise Frozen Text feature (TextNoise), Weak Alignment Loss (WALoss), and Anti-noise Prompt injection. As shown in Table~\ref{tab:ablation1}, introducing WALoss or Anti-Prompt individually yields consistent improvements over the baseline, with gains of +0.06 and +0.19 respectively. Notably, the addition of TextNoise leads to further gains when combined with either WALoss (+0.34) or Anti-Prompt (+0.40). When all three components are enabled, ANPrompt achieves the highest harmonic mean (HM) of 81.70, surpassing the baseline by +0.79. These results verify the complementary nature of each module and their collective effectiveness in enhancing robustness under noise.

\begin{table}[ht]
\centering
\caption{The Effectiveness of different components in ANPrompt. The TextNoise stands for the Weak noise Frozen Text feature.}
\label{tab:ablation1}
\begin{tabular}{ccc|ccc}
\toprule
TextNoise & WALoss & Anti-Prompt  & HM \\  \midrule
\XSolidBrush & \XSolidBrush& \XSolidBrush & 80.91\\ 
\XSolidBrush & \Checkmark & \XSolidBrush  & 80.97\textbf{(+0.06)} \\ 
\XSolidBrush & \XSolidBrush & \Checkmark  & 81.10\textbf{(+0.19)} \\ 
\Checkmark & \Checkmark & \XSolidBrush  & 81.25\textbf{(+0.34)} \\ 
\XSolidBrush & \Checkmark& \Checkmark & 81.27\textbf{(+0.36)} \\ 
\Checkmark & \XSolidBrush & \Checkmark  & 81.31\textbf{(+0.40)} \\ 
\Checkmark & \Checkmark & \Checkmark  & \textbf{81.70}\textbf{(+0.79)} \\ 
\bottomrule
\end{tabular}
\end{table}

\subsubsection{Effect of Prompt Token Length}

We study the effect of varying the number of prompt tokens $T$ on performance. As shown in Table~\ref{tab:length}, increasing $T$ from 1 to 5 gradually improves performance across base and novel classes, leading to a peak harmonic mean (HM) of 81.70\% at $T=5$. This suggests that moderate prompt length enhances generalization. However, performance saturates at $T=6$, where no further gain is observed, which proved the effectiveness of the ablation experiment on token length. Therefore, setting $T=5$ offers the best trade-off between expressiveness and stability.

\begin{table}[h]
\centering
\caption{The length of Prompt Token.}
\label{tab:length}
\begin{tabular}{c|cccccc}
\toprule
$T $ & 1 & 2 & 3 & 4 & 5 & 6 \\
\midrule
$B $ & 85.80 & 86.08 & 86.13 & 86.08 & \textbf{86.15} & 86.09 \\
$N $ & 75.87 & 75.25 & 75.47 & 75.87 & \textbf{77.70} & 77.67 \\
$HM $ & 80.53 & 80.30 & 80.45 & 80.66 & \textbf{81.70} & 81.66 \\
\bottomrule
\end{tabular}
\end{table}

\subsubsection{Effect of weight of Weak Noise Frozen Text feature}

We examine the sensitivity of ANPrompt to the weighting factor applied to the Weak Noise Frozen Text feature. As shown in Table~\ref{tab:weight}, setting the weight to a small value (0.001) yields the best harmonic mean (81.70), indicating effective integration of weak noise signals. As the weight increases, performance on novel classes consistently degrades, leading to a gradual drop in HM. This trend suggests that overemphasizing the weak noise branch introduces distributional bias and harms generalization. These results highlight the necessity of a carefully balanced contribution from TextNoise.

\begin{table}[ht]
\centering
\caption{The Effectiveness of different weight of Weak Noise Frozen Text feature.}
\label{tab:weight}
\begin{tabular}{ccc|c}
\toprule
 Weight & Base & Novel & HM \\  \midrule
 0.001&\textbf{86.15}&\textbf{77.70} & \textbf{81.70} \\ 
 0.01& 86.10& 76.84 & 81.21 \\ 
 0.1 & 86.09& 76.48 & 81.00 \\ 
 1.0 & 86.02& 75.51 & 80.43 \\ 
\bottomrule
\end{tabular}
\end{table} 

\subsubsection{Logits Distance of WALoss}

We conducted an ablation experiment on the consistency constraint of WALoss ($\mathcal{L}_{\text{WA}}$). As shown in Table~\ref{tab:base_new_hm}, we show the performance results when using different consistency constraints (including cosine, L1 and MSE). As you can see, our method ANPrompt has achieved the best results. Specifically, the result of ANPrompt in base is 86.15\%, which is 5.57\% higher than that of traditional MSE. The result on new is 77.70\%, which is 5.82\% higher than the traditional MSE. On the HM, the result of ANPrompt is 81.70\%, which is 1.07\%, 5.72\% and 0.22\% higher than L1, MSE and cosine, respectively. These results show that kl loss is most conducive to learning consistency under the condition of weak semantic perturbation to improve robustness.

\begin{table}[ht]
\centering
\caption{Performance comparison of different calculative methods on WALoss ($\mathcal{L}_{\text{WA}}$).}
\begin{tabular}{lccc}
\toprule
Method & Base & New & HM \\
\midrule
L1     & 86.10 & 75.81 & 80.63 \\
MSE    & 80.58 & 71.88 & 75.98 \\
Cosine    & 85.27 & 78.01 & 81.48 \\
Ours   & \textbf{86.15} & \textbf{77.70} & \textbf{81.70} \\
\bottomrule
\end{tabular}
\label{tab:base_new_hm}
\end{table}

\subsubsection{Ablation study on $\theta$}

As shown in Table~\ref{tab:t3}, the HM score exhibits a consistent trend with increasing values of $\theta$. Specifically, performance gradually improves from $\theta = 0.1$ to $\theta = 0.7$, indicating that the progressive incorporation of alignment signals—despite being accompanied by weak semantic noise—introduces a beneficial level of learning difficulty. This increased difficulty encourages the model to more effectively acquire downstream semantic knowledge and enhance robustness. The performance peak at $\theta = 0.7$ (81.70\%) suggests an optimal trade-off, where the challenge introduced by noise is sufficient to promote generalization without overwhelming the model. However, as $\theta$ increases beyond $0.7$, a performance decline is observed, implying that excessive emphasis on alignment suppresses robustness-related cues and diminishes the advantages conferred by controlled noise. These findings highlight the role of moderate noise as a regularizing signal that facilitates generalization, while excessive alignment leads to suboptimal performance.

\begin{table}[h]
\centering
\caption{Comparison of different number of $\theta$.}
\begin{tabular}{lccc}
\toprule
Method & Base & New & HM \\
\midrule
0.1     & 83.73 & 75.75 & 79.54\\
0.2    & 84.36 & 75.48 & 79.67 \\
0.3    & 84.78 & 75.23& 79.72\\
0.4    &85.17 &76.32 &80.50\\
0.5    &85.67 & 76.48 &80.82\\
0.6    &85.98 &76.93 &81.20 \\
 0.7 & \textbf{86.15} & \textbf{77.70} & \textbf{81.70} \\
0.8   & 86.10& 75.85 &80.65 \\
0.9     &85.69 &75.53 &80.29 \\
1.0 &     84.75   &75.38       &79.79\\
\bottomrule
\end{tabular}
\label{tab:t3}
\end{table}

\subsubsection{Ablation on different generation methods of $\gamma$}

We conduct an ablation study to evaluate different strategies for computing the weight $\gamma$ of the weak alignment loss. Specifically, we compare the following three formulations:
\begin{table}[h]
\centering
\caption{Ablation on different generation methods of $\gamma$.}
\begin{tabular}{lccc}
\toprule
Method & Base & New & HM \\
\midrule
Log    & 86.28& 75.66 &80.62\\
mean     & 68.59 & 60.43 & 64.25 \\
Softmax    & 86.29 & 75.89& 80.76\\
Ours & \textbf{86.15} & \textbf{77.70} & \textbf{81.70} \\

\bottomrule
\end{tabular}
\label{tab:t4}
\end{table}

\begin{itemize}
    \item \textbf{Log}. This strategy applies logarithmic smoothing to the standard deviation to mitigate large gradients. However, the logarithmic transformation compresses variation, reducing sensitivity to changes in prediction confidence:
    \begin{equation}
        \gamma = \frac{1}{\log(\mathrm{std}(\ell_R) + 1.1) \cdot |\ell_R|}
    \end{equation}

    \item \textbf{Mean}. This method normalizes the variance by the mean logit value. However, it is highly unstable when the mean is small, often resulting in erratic scaling:
    \begin{equation}
        \gamma = \frac{1}{\left( \frac{\mathrm{std}(\ell_R)}{\mathrm{mean}(\ell_R) + \epsilon} \right) \cdot |\ell_R|}
    \end{equation}

    \item \textbf{Softmax-Entropy}. This approach uses output entropy as a proxy for uncertainty. While conceptually sound, it is sensitive to class imbalance and tends to saturate under confident predictions:
    \begin{equation}
        \gamma = \frac{1}{\mathcal{H}(\mathrm{softmax}(\ell_R)) \cdot |\ell_R|}
    \end{equation}
    where entropy is defined as:
    \begin{equation}
        \mathcal{H}(p) = -\sum p \log(p)
    \end{equation}

\end{itemize}

As shown in Table~\ref{tab:t4}, our proposed variance-adaptive weighting strategy achieves the highest HM score (81.70\%) among all competing methods, demonstrating a clear advantage in overall performance. Compared to the Log and Mean baselines, the former applies logarithmic smoothing to the standard deviation to suppress gradient fluctuations but inadvertently compresses the model’s responsiveness to semantic variations. The latter normalizes the standard deviation by the mean, yet becomes numerically unstable when the mean approaches zero, often resulting in erratic weight scaling. The Softmax-Entropy method attempts to use output entropy as a proxy for uncertainty. Although conceptually reasonable, it tends to saturate under high-confidence predictions, failing to capture fine-grained semantic perturbations.

In contrast, our approach dynamically adjusts the alignment strength based on the dispersion of the logits, maintaining stability while enhancing the model’s sensitivity to subtle semantic shifts. Moreover, it exhibits reduced sensitivity to class imbalance and low-magnitude outputs, thereby improving its generalization capability. Overall, the results suggest that variance-adaptive weighting strikes a better balance between stability and sensitivity, significantly enhancing the model’s robustness and cross-category generalization under weak semantic perturbations.

\subsubsection{Ablation on different noise injection methods}
\begin{table}[h]
\centering
\caption{Performance comparison of different noise injection.}
\begin{tabular}{lccc}
\toprule
Method & Base & New & HM \\
\midrule
Synreplace & 85.90 & 77.43 & 81.44\\
mask & 85.99 & 77.41 & 81.47 \\
Shuffle & 85.90 & 77.16 & 81.29 \\
Drop &85.97 & 76.94 & 81.21 \\
Ours & \textbf{86.15} & \textbf{77.70} &\textbf{81.70}\\
\bottomrule
\end{tabular}
\label{tab:textnoise}
\end{table}

Table \ref{tab:textnoise} compares the performance of different noise injection methods, evaluated on Base accuracy, New class accuracy, and Harmonic Mean (HM). The methods tested include Synonym Replacement (Synreplace), Masking, Shuffling, Dropping, and our proposed method (Ours). Among the traditional methods, Synonym Replacement achieves the lowest performance, with a New class accuracy of 77.43\% and an HM of 81.44\%. Masking and Shuffling show slightly better results, with New class accuracies of 77.41\% and 77.16\%, and HMs of 81.47\% and 81.29\%, respectively. Dropping yields a New class accuracy of 76.94\% and an HM of 81.21\%. In contrast, our method (Ours) outperforms all others, with a Base accuracy of 86.15\%, a New class accuracy of 77.70\%, and the highest HM of 81.70\%. This highlights the effectiveness of our approach in improving robustness to weak semantic noise while maintaining strong generalization to new categories.nOur method achieves this by integrating weak semantic perturbations into the prompt tokens and enforcing logits-level consistency via Weak Alignment Loss (WALoss). This design allows the model to handle noise-induced variations without sacrificing semantic fidelity, providing superior robustness compared to traditional noise injection techniques.

\subsubsection{Ablation on injection layers}

\begin{table}[h]
\centering
\caption{Ablation on injection layers.}
\begin{tabular}{lccc}
\toprule
Layers & Base & New & HM \\
\midrule
$\{$1-3$\}$ & 85.57 & 72.64 & 78.58 \\
$\{$1-6$\}$ & 85.71 & 73.97 & 79.41 \\
$\{$1-9$\}$ & 85.93 & 74.96 & 80.07 \\
$\{$1-12$\}$ & 86.02 & 73.98 & 79.55 \\
$\{$3-6$\}$ & 85.72 & 75.49 & 80.28 \\
$\{$3-9$\}$ & 85.96 & 75.20 & 80.22 \\
$\{$3-12$\}$ & 85.94 & 75.25 & 80.24 \\
$\{$6-9$\}$ & 85.85 & 76.72 & 81.03 \\
$\{$6-12$\}$ & \textbf{86.15} & \textbf{77.70} & \textbf{81.70} \\
$\{$9-12$\}$ & 85.74 & 75.54 & 80.20 \\
\bottomrule
\end{tabular}
\label{tab:injectionlayers}
\end{table}

As shown in the injection  layers ablation results in Table \ref{tab:injectionlayers}, the layer position at which Anti-noise Prompts are injected has a significant impact on model performance. When prompts are injected into shallower layers (e.g., layers 1–3 or 1–6), the model achieves reasonable performance on Base classes but exhibits weaker generalization on New classes, along with a relatively lower Harmonic Mean (HM). As the injection layers move progressively deeper (e.g., layers 6–9 or 6–12), performance on both Base and New classes improves steadily, reaching its optimum when prompts are injected into layers 6–12, where the HM reaches 81.70\%.

This phenomenon may be attributed to the hierarchical nature of semantic information captured at different depths in vision-language models. Shallow layers typically extract local, low-level visual features, while deeper layers tend to model global, high-level semantic representations. Injecting  Anti-noise Prompts  into deeper layers (e.g., layers 6–12) allows for more effective introduction of mild perturbations at the semantic level, enabling the model to learn robustness to semantic variations in high-level feature spaces. As a result, the model maintains its discriminative ability on Base classes while significantly improving generalization to New classes. In contrast, injecting prompts into shallower layers (e.g., 1–3) may fail to adequately integrate semantic-level perturbations, whereas injection into even deeper layers (e.g., 9–12) could introduce excessive noise interference, leading to performance degradation. Therefore, selecting layers 6–12 as the injection interval achieves an optimal balance between semantic expressiveness and noise robustness.

\subsection{Grad-CAM Visualization}
\begin{figure*}[!htbp]
\centering
\includegraphics[scale=0.5]{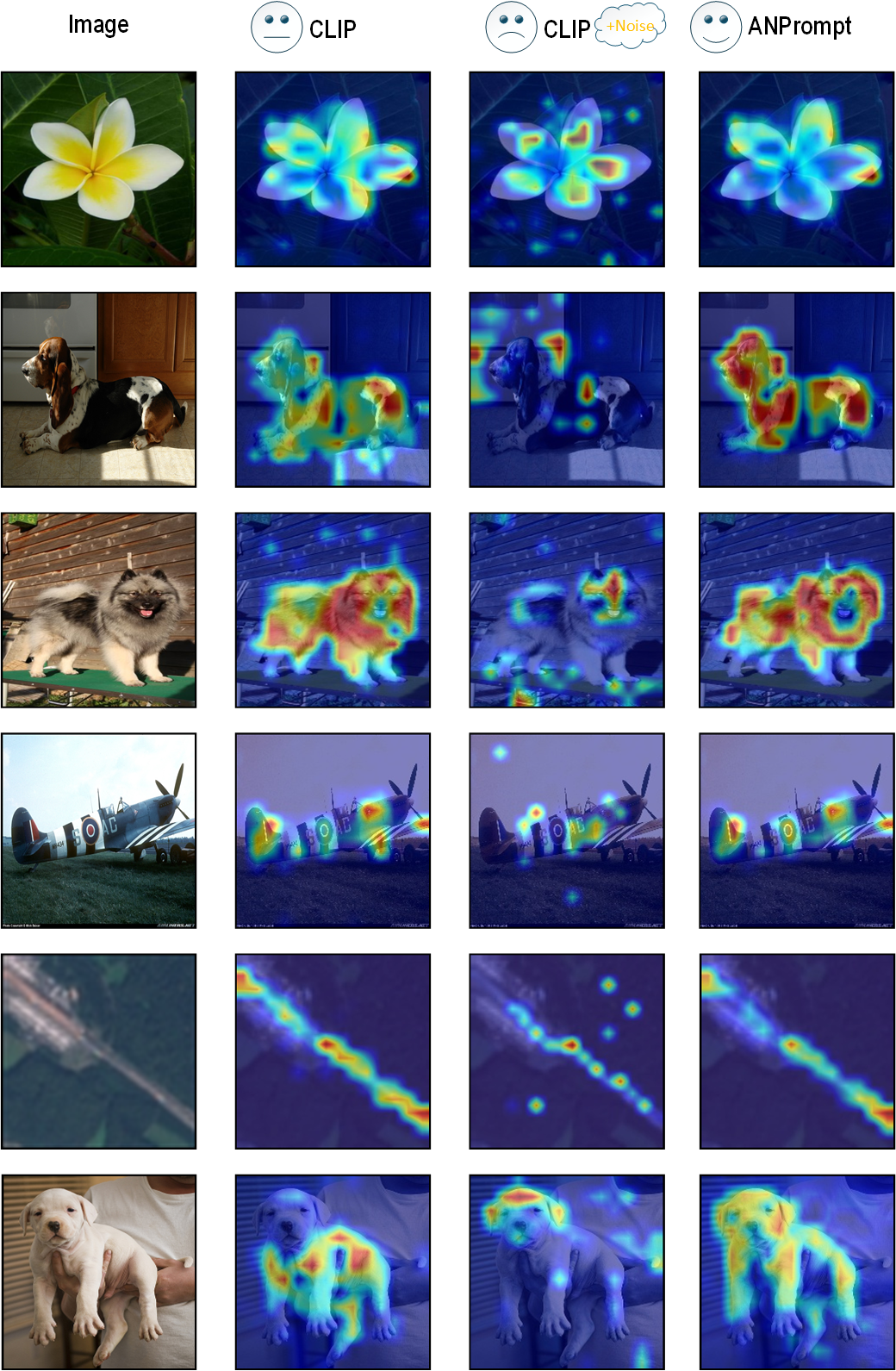}
\caption{ Comparison of Grad-CAM visualizations. From left to right: original image, standard CLIP, CLIP with noisy text, and our ANPrompt with noisy text. The standard CLIP model is susceptible to textual noise, leading to inaccurate and dispersed attention. In contrast, our ANPrompt model effectively mitigates noise interference, maintaining concentrated and semantically precise heatmaps on the target objects.}
\label{fig:cam}
\end{figure*}

To evaluate the noise resistance of our model, we conducted a Grad-CAM~\cite{gradcam} visualization to compare the attention responses of the pre-trained CLIP model and ANPrompt under weak semantic noise. As shown in Figure ~\ref{fig:cam}, when the text input contains weak semantic perturbations, the attention maps of the CLIP model become significantly dispersed, with some focus on background or semantically irrelevant regions, indicating that its visual representations are susceptible to noise interference. In contrast, under the same noisy conditions, our ANPrompt model consistently maintains its attention on the key regions of the target object, demonstrating superior semantic robustness. This capability is notably evident on fine-grained datasets such as EuroSAT and FGVCAircraft, where ANPrompt accurately pinpoints discriminative local features, further validating its stable recognition performance in complex semantic scenarios.

This observed robustness is attributed to the explicitly designed weak semantic noise training mechanism and the Noise-Resistant Visual Prompt Prototype (NRVPP) integrated into the ANPrompt framework. By clustering weak noise features and incorporating them into the prompt representations, our approach exposes the model to and helps it adapt to semantic variations during training, thereby enhancing its inherent resistance to noise. Concurrently, the proposed Weak Alignment Loss (WALoss) imposes a consistency constraint at the logits level, guiding the model to effectively filter out noise during inference and firmly anchor its predictions to the core semantic concepts. Consequently, ANPrompt suppresses semantic drift at the feature level while reinforcing discriminative power at the decision level, ultimately leading to more stable and interpretable visual attention.

\section{CONCLUSION}
\label{conclusion}
We proposed ANPrompt, a robust prompt tuning framework that explicitly models weak semantic perturbations to improve generalization in vision-language models. By injecting weak semantic perturbation into the aligment process and prompt representation, ANPrompt enhances robustness under theweak semantic perturbation. Experimental results on 11 benchmarks demonstrate its clear advantages over existing methods, highlighting the effectiveness of incorporating weak semantic perturbation noise into prompt tuning.

\bibliographystyle{unsrt}
\bibliography{ANPrompt/reference}

\begin{thebibliography}{10}

\bibitem{openworldauc}
Cong Hua, Qianqian Xu, Zhiyong Yang, Zitai Wang, Shilong Bao, and Qingming Huang.
\newblock Openworldauc: Towards unified evaluation and optimization for open-world prompt tuning.
\newblock {\em arXiv preprint arXiv:2505.05180}, 2025.

\bibitem{openvclip}
Zejia Weng, Xitong Yang, Ang Li, Zuxuan Wu, and Yu-Gang Jiang.
\newblock Open-vclip: Transforming clip to an open-vocabulary video model via interpolated weight optimization.
\newblock In {\em International conference on machine learning}, pages 36978--36989. PMLR, 2023.

\bibitem{xu2024llava}
Guowei Xu, Peng Jin, Ziang Wu, Hao Li, Yibing Song, Lichao Sun, and Li~Yuan.
\newblock Llava-cot: Let vision language models reason step-by-step.
\newblock {\em arXiv preprint arXiv:2411.10440}, 2024.

\bibitem{awt}
Yuhan Zhu, Yuyang Ji, Zhiyu Zhao, Gangshan Wu, and Limin Wang.
\newblock Awt: Transferring vision-language models via augmentation, weighting, and transportation.
\newblock {\em Advances in Neural Information Processing Systems}, 37:25561--25591, 2024.

\bibitem{gallop}
Marc Lafon, Elias Ramzi, Cl{\'e}ment Rambour, Nicolas Audebert, and Nicolas Thome.
\newblock Gallop: Learning global and local prompts for vision-language models.
\newblock In {\em European Conference on Computer Vision}, pages 264--282. Springer, 2024.

\bibitem{qnet}
Boya Shi, Zhengqin Xu, Shuai Jia, and Chao Ma.
\newblock Prompt learning with quaternion networks.
\newblock In {\em The Twelfth International Conference on Learning Representations}, 2024.

\bibitem{coop}
Kaiyang Zhou, Jingkang Yang, Chen~Change Loy, and Ziwei Liu.
\newblock Learning to prompt for vision-language models.
\newblock In {\em International Journal of Computer Vision}, pages 2337--2348, 2022.

\bibitem{cocoop}
Kaiyang Zhou, Jingkang Yang, Chen~Change Loy, and Ziwei Liu.
\newblock Conditional prompt learning for vision-language models.
\newblock In {\em Computer Vision and Pattern Recognition}, pages 16795--16804, 2022.

\bibitem{kgcoop}
Hantao Yao, Rui Zhang, and Changsheng Xu.
\newblock Visual-language prompt tuning with knowledge-guided context optimization.
\newblock In {\em Proceedings of the IEEE/CVF conference on computer vision and pattern recognition}, pages 6757--6767, 2023.

\bibitem{maple}
Muhammad~Uzair Khattak, Hanoona Rasheed, Muhammad Maaz, Salman Khan, and Fahad~Shahbaz Khan.
\newblock Maple: Multi-modal prompt learning.
\newblock In {\em Computer Vision and Pattern Recognition}, pages 19113--19122, 2023.

\bibitem{mmrl}
Yuncheng Guo and Xiaodong Gu.
\newblock Mmrl: Multi-modal representation learning for vision-language models.
\newblock In {\em Proceedings of the Computer Vision and Pattern Recognition Conference}, pages 25015--25025, 2025.

\bibitem{dapt}
Fei Zhang, Tianfei Zhou, Jiangchao Yao, Ya~Zhang, Ivor~W Tsang, and Yanfeng Wang.
\newblock Decouple before align: Visual disentanglement enhances prompt tuning.
\newblock {\em IEEE Transactions on Pattern Analysis and Machine Intelligence}, 2025.

\bibitem{argue}
Xinyu Tian, Shu Zou, Zhaoyuan Yang, and Jing Zhang.
\newblock Argue: Attribute-guided prompt tuning for vision-language models.
\newblock In {\em Proceedings of the IEEE/CVF Conference on Computer Vision and Pattern Recognition}, pages 28578--28587, 2024.

\bibitem{noise1}
Jiasen Lu, Dhruv Batra, Devi Parikh, and Stefan Lee.
\newblock Vilbert: Pretraining task-agnostic visiolinguistic representations for vision-and-language tasks.
\newblock {\em Advances in neural information processing systems}, 32, 2019.

\bibitem{l1}
Hao Li, He~Cao, Bin Feng, Yanjun Shao, Xiangru Tang, Zhiyuan Yan, Li~Yuan, Yonghong Tian, and Yu~Li.
\newblock Beyond chemical qa: Evaluating llm's chemical reasoning with modular chemical operations.
\newblock {\em arXiv preprint arXiv:2505.21318}, 2025.

\bibitem{l2}
Hao Li, Jinfa Huang, Peng Jin, Guoli Song, Qi~Wu, and Jie Chen.
\newblock Weakly-supervised 3d spatial reasoning for text-based visual question answering.
\newblock {\em IEEE Transactions on Image Processing}, 32:3367--3382, 2023.

\bibitem{l3}
Hao Li, Yanhao Jia, Peng Jin, Zesen Cheng, Kehan Li, Jialu Sui, Chang Liu, and Li~Yuan.
\newblock Freestyleret: retrieving images from style-diversified queries.
\newblock In {\em European Conference on Computer Vision}, pages 258--274. Springer, 2024.

\bibitem{l4}
Hao Li, Xu~Li, Belhal Karimi, Jie Chen, and Mingming Sun.
\newblock Joint learning of object graph and relation graph for visual question answering.
\newblock In {\em 2022 IEEE International Conference on Multimedia and Expo (ICME)}, pages 01--06. IEEE, 2022.

\bibitem{l5}
Hao Li, Da~Long, Li~Yuan, Yu~Wang, Yonghong Tian, Xinchang Wang, and Fanyang Mo.
\newblock Decoupled peak property learning for efficient and interpretable electronic circular dichroism spectrum prediction.
\newblock {\em Nature Computational Science}, 5(3):234--244, 2025.

\bibitem{clip}
Alec Radford, Jong~Wook Kim, Chris Hallacy, Aditya Ramesh, Gabriel Goh, Sandhini Agarwal, Girish Sastry, Amanda Askell, Pamela Mishkin, Jack Clark, et~al.
\newblock Learning transferable visual models from natural language supervision.
\newblock In {\em International Conference on Machine Learning}, pages 8748--8763, 2021.

\bibitem{align}
Chao Jia, Yinfei Yang, Ye~Xia, Yi-Ting Chen, Zarana Parekh, Hieu Pham, Quoc Le, Yun-Hsuan Sung, Zhen Li, and Tom Duerig.
\newblock Scaling up visual and vision-language representation learning with noisy text supervision.
\newblock In {\em International conference on machine learning}, pages 4904--4916. PMLR, 2021.

\bibitem{blip}
Junnan Li, Dongxu Li, Caiming Xiong, and Steven Hoi.
\newblock Blip: Bootstrapping language-image pre-training for unified vision-language understanding and generation.
\newblock In {\em International conference on machine learning}, pages 12888--12900. PMLR, 2022.

\bibitem{filip}
Lewei Yao, Runhui Huang, Lu~Hou, Guansong Lu, Minzhe Niu, Hang Xu, Xiaodan Liang, Zhenguo Li, Xin Jiang, and Chunjing Xu.
\newblock Filip: Fine-grained interactive language-image pre-training.
\newblock {\em arXiv preprint arXiv:2111.07783}, 2021.

\bibitem{florence}
L~Yuan, D~Chen, YL~Chen, N~Codella, X~Dai, J~Gao, H~Hu, X~Huang, B~Li, C~Li, et~al.
\newblock Florence: A new foundation model for computer vision. arxiv 2021.
\newblock {\em arXiv preprint arXiv:2111.11432}, 2021.

\bibitem{lit}
Xiaohua Zhai, Xiao Wang, Basil Mustafa, Andreas Steiner, Daniel Keysers, Alexander Kolesnikov, and Lucas Beyer.
\newblock Lit: Zero-shot transfer with locked-image text tuning.
\newblock In {\em Proceedings of the IEEE/CVF conference on computer vision and pattern recognition}, pages 18123--18133, 2022.

\bibitem{li2024freestyleret}
Hao Li, Yanhao Jia, Peng Jin, Zesen Cheng, Kehan Li, Jialu Sui, Chang Liu, and Li~Yuan.
\newblock Freestyleret: retrieving images from style-diversified queries.
\newblock In {\em European Conference on Computer Vision}, pages 258--274. Springer, 2024.

\bibitem{NLIP}
Runhui Huang, Yanxin Long, Jianhua Han, Hang Xu, Xiwen Liang, Chunjing Xu, and Xiaodan Liang.
\newblock Nlip: Noise-robust language-image pre-training.
\newblock In {\em Proceedings of the AAAI Conference on Artificial Intelligence}, volume~37, pages 926--934, 2023.

\bibitem{li2025beyond}
Hao Li, He~Cao, Bin Feng, Yanjun Shao, Xiangru Tang, Zhiyuan Yan, Li~Yuan, Yonghong Tian, and Yu~Li.
\newblock Beyond chemical qa: Evaluating llm's chemical reasoning with modular chemical operations.
\newblock {\em arXiv preprint arXiv:2505.21318}, 2025.

\bibitem{li2023weakly}
Hao Li, Jinfa Huang, Peng Jin, Guoli Song, Qi~Wu, and Jie Chen.
\newblock Weakly-supervised 3d spatial reasoning for text-based visual question answering.
\newblock {\em IEEE Transactions on Image Processing}, 32:3367--3382, 2023.

\bibitem{pcl}
Shuo Li, Fang Liu, Licheng Jiac, Lingling Li, Puhua Chen, Xu~Liu, and Wenping Ma.
\newblock Prompt-based concept learning for few-shot class-incremental learning.
\newblock {\em IEEE Transactions on Circuits and Systems for Video Technology}, 2025.

\bibitem{hipl}
Xiaotian Yin, Jiamin Wu, Wenfei Yang, Xu~Zhou, Shifeng Zhang, and Tianzhu Zhang.
\newblock Hierarchy-aware interactive prompt learning for few-shot classification.
\newblock {\em IEEE Transactions on Circuits and Systems for Video Technology}, 2024.

\bibitem{sod}
Kunpeng Wang, Zhengzheng Tu, Chenglong Li, Zhengyi Liu, and Bin Luo.
\newblock Unified-modal salient object detection via adaptive prompt learning.
\newblock {\em IEEE Transactions on Circuits and Systems for Video Technology}, 2025.

\bibitem{clipadapter}
Peng Gao, Shijie Geng, Renrui Zhang, Teli Ma, Rongyao Fang, Yongfeng Zhang, Hongsheng Li, and Yu~Qiao.
\newblock Clip-adapter: Better vision-language models with feature adapters.
\newblock {\em International Journal of Computer Vision}, 132(2):581--595, 2024.

\bibitem{mma}
Lingxiao Yang, Ru-Yuan Zhang, Yanchen Wang, and Xiaohua Xie.
\newblock Mma: Multi-modal adapter for vision-language models.
\newblock In {\em Proceedings of the IEEE/CVF Conference on Computer Vision and Pattern Recognition}, pages 23826--23837, 2024.

\bibitem{metaadapter}
Lin Song, Ruoyi Xue, Hang Wang, Hongbin Sun, Yixiao Ge, Ying Shan, et~al.
\newblock Meta-adapter: An online few-shot learner for vision-language model.
\newblock {\em Advances in Neural Information Processing Systems}, 36:55361--55374, 2023.

\bibitem{protext}
Muhammad~Uzair Khattak, Muhammad~Ferjad Naeem, Muzammal Naseer, Luc Van~Gool, and Federico Tombari.
\newblock Learning to prompt with text only supervision for vision-language models.
\newblock {\em arXiv preprint arXiv:2401.02418}, 2024.

\bibitem{cpr}
Haoxing Chen, Yaohui Li, Zizheng Huang, Yan Hong, Zhuoer Xu, Zhangxuan Gu, Jun Lan, Huijia Zhu, and Weiqiang Wang.
\newblock Conditional prototype rectification prompt learning.
\newblock {\em IEEE Transactions on Circuits and Systems for Video Technology}, 2025.

\bibitem{pmp}
Jun Liu, Ziqian Lu, Hao Luo, Zheming Lu, and Yangming Zheng.
\newblock Progressive multi-prompt learning for vision-language models.
\newblock {\em IEEE Transactions on Circuits and Systems for Video Technology}, 2025.

\bibitem{promptsrc}
Muhammad~Uzair Khattak, Syed~Talal Wasim, Muzammal Naseer, Salman Khan, Ming-Hsuan Yang, and Fahad~Shahbaz Khan.
\newblock Self-regulating prompts: Foundational model adaptation without forgetting.
\newblock In {\em Proceedings of the IEEE/CVF international conference on computer vision}, pages 15190--15200, 2023.

\bibitem{prograd}
Beier Zhu, Yulei Niu, Yucheng Han, Yue Wu, and Hanwang Zhang.
\newblock Prompt-aligned gradient for prompt tuning.
\newblock In {\em Proceedings of the IEEE/CVF international conference on computer vision}, pages 15659--15669, 2023.

\bibitem{hpt}
Yubin Wang, Xinyang Jiang, De~Cheng, Dongsheng Li, and Cairong Zhao.
\newblock Learning hierarchical prompt with structured linguistic knowledge for vision-language models.
\newblock In {\em Proceedings of the AAAI conference on artificial intelligence}, volume~38, pages 5749--5757, 2024.

\bibitem{tcp}
Hantao Yao, Rui Zhang, and Changsheng Xu.
\newblock Tcp: Textual-based class-aware prompt tuning for visual-language model.
\newblock In {\em Proceedings of the IEEE/CVF Conference on Computer Vision and Pattern Recognition}, pages 23438--23448, 2024.

\bibitem{clipkd}
Chuanguang Yang, Zhulin An, Libo Huang, Junyu Bi, Xinqiang Yu, Han Yang, Boyu Diao, and Yongjun Xu.
\newblock Clip-kd: An empirical study of clip model distillation.
\newblock In {\em Proceedings of the IEEE/CVF Conference on Computer Vision and Pattern Recognition}, pages 15952--15962, 2024.

\bibitem{comkdclip}
Yifan Chen, Xiaozhen Qiao, Zhe Sun, and Xuelong Li.
\newblock Comkd-clip: Comprehensive knowledge distillation for contrastive language-image pre-traning model.
\newblock {\em arXiv preprint arXiv:2408.04145}, 2024.

\bibitem{promptkd}
Zheng Li, Xiang Li, Xinyi Fu, Xin Zhang, Weiqiang Wang, Shuo Chen, and Jian Yang.
\newblock Promptkd: Unsupervised prompt distillation for vision-language models.
\newblock In {\em Proceedings of the IEEE/CVF Conference on Computer Vision and Pattern Recognition}, pages 26617--26626, 2024.

\bibitem{caspl}
Ge~Wu, Xin Zhang, Zheng Li, Zhaowei Chen, Jiajun Liang, Jian Yang, and Xiang Li.
\newblock Cascade prompt learning for vision-language model adaptation.
\newblock In {\em European Conference on Computer Vision}, pages 304--321. Springer, 2024.

\bibitem{atprompt}
Zheng Li, Yibing Song, Ming-Ming Cheng, Xiang Li, and Jian Yang.
\newblock Advancing textual prompt learning with anchored attributes.
\newblock {\em arXiv preprint arXiv:2412.09442}, 2024.

\bibitem{dpc}
Haoyang Li, Liang Wang, Chao Wang, Jing Jiang, Yan Peng, and Guodong Long.
\newblock Dpc: Dual-prompt collaboration for tuning vision-language models.
\newblock In {\em Proceedings of the Computer Vision and Pattern Recognition Conference}, pages 25623--25632, 2025.

\bibitem{dept}
Ji~Zhang, Shihan Wu, Lianli Gao, Heng~Tao Shen, and Jingkuan Song.
\newblock Dept: Decoupled prompt tuning.
\newblock In {\em Proceedings of the IEEE/CVF Conference on Computer Vision and Pattern Recognition}, pages 12924--12933, 2024.

\bibitem{2sfs}
Matteo Farina, Massimiliano Mancini, Giovanni Iacca, and Elisa Ricci.
\newblock Rethinking few-shot adaptation of vision-language models in two stages.
\newblock {\em arXiv preprint arXiv:2503.11609}, 2025.

\bibitem{coprompt}
Shuvendu Roy and Ali Etemad.
\newblock Consistency-guided prompt learning for vision-language models.
\newblock In {\em International Conference on Learning Representations}, 2024.

\bibitem{tac}
Fusheng Hao, Fengxiang He, Fuxiang Wu, Tichao Wang, Chengqun Song, and Jun Cheng.
\newblock Task-aware clustering for prompting vision-language models.
\newblock In {\em Proceedings of the Computer Vision and Pattern Recognition Conference}, pages 14745--14755, 2025.

\bibitem{nlprompt}
Bikang Pan, Qun Li, Xiaoying Tang, Wei Huang, Zhen Fang, Feng Liu, Jingya Wang, Jingyi Yu, and Ye~Shi.
\newblock Nlprompt: Noise-label prompt learning for vision-language models.
\newblock In {\em Proceedings of the Computer Vision and Pattern Recognition Conference}, pages 19963--19973, 2025.

\bibitem{joapr}
Yuncheng Guo and Xiaodong Gu.
\newblock Joapr: Cleaning the lens of prompt learning for vision-language models.
\newblock In {\em Proceedings of the IEEE/CVF Conference on Computer Vision and Pattern Recognition}, pages 28695--28705, 2024.

\bibitem{clipcleaner}
Chen Feng, Georgios Tzimiropoulos, and Ioannis Patras.
\newblock Clipcleaner: Cleaning noisy labels with clip.
\newblock In {\em Proceedings of the 32nd ACM International Conference on Multimedia}, pages 876--885, 2024.

\bibitem{caltech101}
Li~Fei-Fei, Rob Fergus, and Pietro Perona.
\newblock Learning generative visual models from few training examples: An incremental bayesian approach tested on 101 object categories.
\newblock In {\em Computer Vision and Pattern Recognition workshop}, pages 178--178, 2004.

\bibitem{oxfordpets}
Omkar~M Parkhi, Andrea Vedaldi, Andrew Zisserman, and CV~Jawahar.
\newblock Cats and dogs.
\newblock In {\em Computer Vision and Pattern Recognition}, pages 3498--3505, 2012.

\bibitem{Flowers102}
Maria-Elena Nilsback and Andrew Zisserman.
\newblock Automated flower classification over a large number of classes.
\newblock In {\em Sixth Indian Conference on Computer Vision}, pages 722--729, 2008.

\bibitem{food101}
Lukas Bossard, Matthieu Guillaumin, and Luc~Van Gool.
\newblock Food-101 - mining discriminative components with random forests.
\newblock In {\em European Conference on Computer Vision}, pages 446--461, 2014.

\bibitem{StanfordCars}
Jonathan Krause, Michael Stark, Jia Deng, and Li~Fei-Fei.
\newblock 3d object representations for fine-grained categorization.
\newblock In {\em International Conference on Computer Vision workshops}, pages 554--561, 2013.

\bibitem{fgvcaircraft}
Subhransu Maji, Esa Rahtu, Juho Kannala, Matthew Blaschko, and Andrea Vedaldi.
\newblock Fine-grained visual classification of aircraft.
\newblock In {\em arXiv preprint arXiv:1306.5151}, 2013.

\bibitem{Eurosat}
Patrick Helber, Benjamin Bischke, Andreas Dengel, and Damian Borth.
\newblock Eurosat: A novel dataset and deep learning benchmark for land use and land cover classification.
\newblock {\em IEEE Journal of Selected Topics in Applied Earth Observations and Remote Sensing}, 12(7):2217--2226, 2019.

\bibitem{ucf101}
Will Kay, Joao Carreira, Karen Simonyan, Brian Zhang, Chloe Hillier, Sudheendra Vijayanarasimhan, Fabio Viola, Tim Green, Trevor Back, Paul Natsev, et~al.
\newblock Ucf101: A dataset of 101 human actions classes from videos in the wild.
\newblock In {\em arXiv preprint arXiv:1212.0402}, 2017.

\bibitem{dtd}
Mircea Cimpoi, Subhransu Maji, Iasonas Kokkinos, Sammy Mohamed, and Andrea Vedaldi.
\newblock Describing textures in the wild.
\newblock In {\em Computer Vision and Pattern Recognition}, pages 3606--3613, 2014.

\bibitem{sun}
Jianxiong Xiao, James Hays, Krista~A Ehinger, Aude Oliva, and Antonio Torralba.
\newblock Sun database: Large-scale scene recognition from abbey to zoo.
\newblock In {\em Computer Vision and Pattern Recognition}, pages 3485--3492, 2010.

\bibitem{imagenet}
Jia Deng, Wei Dong, Richard Socher, Li-Jia Li, Kai Li, and Li~Fei-Fei.
\newblock Imagenet: A large-scale hierarchical image database.
\newblock In {\em Computer Vision and Pattern Recognition}, pages 248--255, 2009.

\bibitem{imagenetv2}
B.~Recht, R.~Roelofs, L.~Schmidt, and V.~Shankar.
\newblock Do imagenet classifiers generalize to imagenet?
\newblock In {\em International Conference on Machine Learning}, pages 5389--5400, 2019.

\bibitem{imagenets}
H.~Wang, S.~Ge, Z.~Lipton, and E.~P. Xing.
\newblock Learning robust global representations by penalizing local predictive power.
\newblock {\em Neural Information Processing Systems}, 32:1--13, 2019.

\bibitem{imageneta}
D.~Hendrycks, K.~Zhao, S.~Basart, J.~Steinhardt, and D.~Song.
\newblock Natural adversarial examples.
\newblock In {\em Computer Vision and Pattern Recognition}, pages 15262--15271, 2021.

\bibitem{imagenetr}
D.~Hendrycks, S.~Basart, N.~Mu, S.~Kadavath, F.~Wang, E.~Dorundo, R.~Desai, T.~Zhu, S.~Parajuli, M.~Guo, et~al.
\newblock The many faces of robustness: A critical analysis of out-of-distribution generalization.
\newblock In {\em International Conference on Computer Vision}, pages 8340--8349, 2021.

\bibitem{textrefiner}
Jingjing Xie, Yuxin Zhang, Jun Peng, Zhaohong Huang, and Liujuan Cao.
\newblock Textrefiner: Internal visual feature as efficient refiner for vision-language models prompt tuning.
\newblock In {\em Proceedings of the AAAI Conference on Artificial Intelligence}, volume~39, pages 8718--8726, 2025.

\bibitem{surpl}
Liangchen Liu, Nannan Wang, Xi~Yang, Xinbo Gao, and Tongliang Liu.
\newblock Surrogate prompt learning: Towards efficient and diverse prompt learning for vision-language models.
\newblock In {\em Forty-second International Conference on Machine Learning}.

\bibitem{lweib}
Lingxiao Yang, Ru-Yuan Zhang, Qi~Chen, and Xiaohua Xie.
\newblock Learning with enriched inductive biases for vision-language models.
\newblock {\em International Journal of Computer Vision}, 133(6):3746--3761, 2025.

\bibitem{bip}
Hantao Yao, Rui Zhang, Huaihai Lyu, Yongdong Zhang, and Changsheng Xu.
\newblock Bi-modality individual-aware prompt tuning for visual-language model.
\newblock {\em IEEE Transactions on Pattern Analysis and Machine Intelligence}, 2025.

\bibitem{gradcam}
Ramprasaath~R Selvaraju, Michael Cogswell, Abhishek Das, Ramakrishna Vedantam, Devi Parikh, and Dhruv Batra.
\newblock Grad-cam: Visual explanations from deep networks via gradient-based localization.
\newblock In {\em Proceedings of the IEEE international conference on computer vision}, pages 618--626, 2017.

\end{thebibliography}

\end{document}